\documentclass[journal]{IEEEtran}

\usepackage{amsthm,amssymb,graphicx,amsmath,color,amsfonts}
\usepackage[update,prepend]{epstopdf}
\usepackage[noadjust]{cite}
\usepackage[latin1]{inputenc}
\usepackage{pdfpages}
\usepackage{tabulary}
\usepackage{multirow}
\usepackage{bbm} 					% for \mathbbm{1}
\usepackage{comment}
\usepackage{dsfont}					% Fix type 3 fonts
\usepackage{subfigure}
\def\chb#1{{\color{blue}#1}}

\def\nba{{\mathbf{a}}}
\def\nbb{{\mathbf{b}}}

\def\nbd{{\mathbf{d}}}

\def\nbm{{\mathbf{m}}}

\def\nbr{{\mathbf{r}}}

\def\nbx{{\mathbf{x}}}
\def\nby{{\mathbf{y}}}

\def\nbzero{{\mathbf{0}}}
\def\nb1{{\mathbbm{1}}}

\def\nbM{{\mathbf{M}}}

\def\nbR{{\mathbf{R}}}

\def\ncalA{{\mathcal{A}}}
\def\ncalB{{\mathcal{B}}}
\def\ncalC{{\mathcal{C}}}

\def\ncalI{{\mathcal{I}}}

\def\ncalS{{\mathcal{S}}}

\def\ncalX{{\mathcal{X}}}

\def\nbbE{{\mathbb{E}}}

\def\nbbM{{\mathbb{M}}}
\def\nbbN{{\mathbb{N}}}

\def\nbbP{{\mathbb{P}}}

\def\nbbR{{\mathbb{R}}}

\def\nrmd{{\rm d}}

\newtheorem{lemma}{Lemma}

\newtheorem{definition}{Definition}

\newtheorem{theorem}{Theorem}

\def\argmin{\operatorname{arg~min}}

\def\figref#1{Fig.\,\ref{#1}}%

\allowdisplaybreaks % Allows breaking of eqnarray over multiple pages (avoids unnecessary blanks in the document before eqnarray)

\begin{document}
\graphicspath{{./Figures/}}

\title{
%Foundations of Vision-based Localization: An Analytical Framework and Localizability Analysis
%A New Statistical Framework for Modeling and Performance Analysis of Vision-based Localization
A New Statistical Approach to the Performance Analysis of Vision-based Localization
}
% [July 14, 2023] This title is appropriate.

\author{
%Author 1, Author 2, Author 3
Haozhou Hu,~\IEEEmembership{Student Member,~IEEE}, Harpreet S. Dhillon,~\IEEEmembership{Fellow,~IEEE}, R. Michael Buehrer,~\IEEEmembership{Fellow,~IEEE}
\thanks{Preliminary results from this work were presented at WiOpt Workshops 2023~\cite{conf1}.
The authors are with Wireless@VT, Bradley Department of Electrical and Computer Engineering, Virginia Tech, Blacksburg, VA, 24061, USA. Email: \{haozhouhu, hdhillon, rbuehrer\}@vt.edu. The support of the US NSF (Grants CNS-2107276 and CNS-2225511) is gratefully acknowledged.} 
}

\maketitle

\begin{abstract} 
\begin{comment}
\chb{Vision-based positioning is enabled by modern wireless devices equipped with various vision sensors, such as cameras, radar, and Light Detection and Ranging (LiDAR).
These sensors capture landmarks and measure distances to them. 
However, it is challenging to distinguish between landmarks with similar appearances, which complicates the process of finding the target's location.
This work proposes a framework that localizes the target by utilizing the identification and relative positions of multiple generic landmarks.
Landmarks are categorized by appearance into different types and are modeled as a marked Poisson point process (PPP).
Ranges to observed landmarks are measured to construct geometric constraints, which then aid in identifying the combination of observed landmarks.
We prove that the observed landmark combination can almost surely be identified with three noise-free range measurements.
When measurements are noisy, we provide a mathematical characterization of the probability of correctly identifying the landmark combination based on a novel joint distribution of key random variables.
Our results demonstrate that the landmark combination can be identified using ranges, even when individual landmarks are visually indistinguishable. 
This highlights the effectiveness of our approach in addressing the challenge of visual ambiguity in vision-based positioning systems.}
\end{comment}
Many modern wireless devices with accurate positioning needs also have access to vision sensors, such as a camera, radar, and Light Detection and Ranging (LiDAR). 
In scenarios where wireless-based positioning is either inaccurate or unavailable, using information from vision sensors becomes highly desirable for determining the precise location of the wireless device.
Specifically, vision data can be used to estimate distances between the target (where the sensors are mounted) and nearby landmarks. 
However, a significant challenge in positioning using these measurements is the inability to uniquely identify which specific landmark is visible in the data.
For instance, when the target is located close to a lamppost, it becomes challenging to precisely identify the specific lamppost (among several in the region) that is near the target. 
This work proposes a new framework for target localization using range measurements to multiple proximate landmarks. The geometric constraints introduced by these measurements are utilized to narrow down candidate landmark combinations corresponding to the range measurements and, consequently, the target's location on a map. 
By modeling landmarks as a marked Poisson point process (PPP), we show that three noise-free range measurements are sufficient to uniquely determine the correct combination of landmarks in a two-dimensional plane. For noisy measurements, we provide a mathematical characterization of the probability of correctly identifying the observed landmark combination based on a novel joint distribution of key random variables. Our results demonstrate that the landmark combination can be identified using ranges, even when individual landmarks are visually indistinguishable.
\begin{comment}
\chb{Many modern wireless devices with accurate positioning needs also have access to vision sensors, such as a camera, radar, and Light Detection and Ranging (LiDAR). 
When wireless positioning is unreliable or unavailable, vision sensors provide a viable alternative for accurate localization by estimating distances between the target and nearby landmarks. However, uniquely identifying the specific landmark visible in the data poses a significant challenge, for instance, identifying a specific lamppost near the target from several similar ones.
This work proposes a new framework for localization using range measurements to multiple landmarks. 
The geometric constraints introduced by these measurements are utilized to narrow down candidate landmark combinations corresponding to the range measurements and, consequently, the target's location on a map. 
By modeling landmarks as a marked Poisson point process (PPP), we show that three noise-free range measurements are sufficient to uniquely determine the correct combination of landmarks in a two-dimensional plane. 
For noisy measurements, we provide a mathematical characterization of the probability of correctly identifying the observed landmark combination based on a novel joint distribution of key random variables. Our results demonstrate that the landmark combination can be identified using ranges, even when individual landmarks are visually indistinguishable.}
\end{comment}

\end{abstract}


\section{Introduction} \label{sec:intro}
Positioning is fundamental to various location-aware applications, including robotics, vehicle navigation, and asset tracking.
Range-based positioning is a widely used approach in which the target estimates its location by taking distance measurements from reference points with known locations. 
For example, in {\em wireless-based positioning}, the distances from the target to wireless nodes (known as anchors) are estimated through wireless transmissions, such as time-of-flight or signal strength measurements. In contrast, in {\em vision-based positioning}, the target's location relative to reference points (known as landmarks) is estimated using visual information captured by cameras or other imaging sensors.
The landmarks are detected in the visual data, and their known locations are used to calculate the target's position through geometric relationships.
The target's global location is usually estimated in two steps: (1) determining the relative locations to reference points and (2) mapping the relative locations to the global location. 
While the foundations of the first step are well-understood in both wireless-based and vision-based positioning methods, the second step in vision-based positioning presents unique challenges.
The mathematical foundation for this mapping remains largely unexplored because of a key challenge: landmarks in vision-based systems can appear visually similar, making it difficult to distinguish between them.
This similarity can lead to ambiguities when localizing the target, as different locations may produce similar measurements. Thus, further mathematical investigation is required to address these ambiguities in vision-based positioning systems rigorously, which is the focus of this paper. 
In particular, we model landmarks as marked points on a map, where points with the same mark are visually indistinguishable. Then using tools from stochastic geometry, we develop a statistical approach that allows us to rigorously characterize the probability of correctly identifying the observed landmark combination based on just the noisy range measurements from the vision data.

\begin{comment}
\chb{In this paper, we model landmarks as marked points, where points with the same mark are visually indistinguishable. Geometric constraints are derived from the noisy range measurements to visible landmarks and are then used to identify the landmark combination observed by the target.
To evaluate localization performance, we analytically derive the probability of correctly identifying the observed landmark combination, which indicates the effectiveness of our approach in differentiating among similar landmarks despite measurement noise. }
\end{comment}

\subsection{Related Work} 
Although classical algorithms for wireless-based positioning are not directly applied in our work, they provide valuable insights due to the methodological similarities between wireless-based positioning and vision-based positioning.
The usual approach in wireless-based positioning typically involves an estimation-theoretic formulation, which often depends on several characteristics of the wireless links between the target and the anchors, including received signal strength, time of arrival, time difference of arrival, and direction of arrival~\cite{wireless}.  
This research area is grounded in solid mathematical principles and has been explored from various perspectives, including fundamental positioning bounds~\cite{crlb1, crlb2, crlb3}, positioning algorithms, and their performance in different propagation environments~\cite{del2017survey}. 
For a comprehensive and accessible overview of this topic, readers can refer to~\cite{buehrerhandbook}.
%Another line of work  Feature matching: RSSI CSI
%the channel state information~\cite{csi}

Compared to wireless-based positioning, vision-based positioning relies on vision sensors to localize the target without requiring infrastructure such as base stations, access points, or satellites.
There are two main streams in this research area: (i) Image localization, which focuses on tagging photos with geographical information, including the locations where they were taken, and (ii) Simultaneous Localization and Mapping (SLAM), which tracks the target's movement while simultaneously constructing or updating the map of the environment.
Image localization is based on the idea that the image captures the surroundings of the target location.
By analyzing the image's content, the location information can be inferred. 
For example, we can infer the location where a photo was taken by identifying recognizable landmarks in the image and mapping them to their known positions, such as associating the Eiffel Tower with its geographical coordinates.
In~\cite{city-scale, visual-place, lin2013cross}, the street view and user-shared photos from social networks are used to create a database for localization. 
These images contain rich annotations, metadata, and location information. 
To determine the unknown location where the image was obtained, they extract features and compare them with images in the database.
The similarities between the query image and reference images in the database are then calculated, and the location of the query image is estimated based on the locations of the most similar reference images.
Some work~\cite{singh2016semantically, crandall2016recognizing} incorporates semantic information from images, such as texts, architectural style, and urban structures. 
These features are extracted and utilized within the {\em bag of words} model~\cite{csurka2004visual}. 
The localization accuracy is further improved with the use of more detailed maps~\cite{li2016worldwide, saurer2016image}, including topography labels and 3-dimensional models. 
In~\cite{ground2aerial}, the landmarks are modeled as various types of nodes, and the geometric placements of these landmarks are used to localize the target. 
The graph of landmarks, which represents their adjacency relationships, is utilized to identify potential locations for the target on the map.
Work done recently on image localization use deep learning. 
A variety of deep neural networks have been proposed for localization, to name a few: PoseNet~\cite{kendall2015posenet}, MapNet~\cite{geometry-aware}, CamNet~\cite{ding2019camnet} and LTSMs~\cite{walch2017image}.
They use public image datasets, such as 7-Scenes~\cite{7scene} and the Oxford Robotcar Dataset, which consist of images captured within a finite region, to train neural networks and evaluate localization accuracy.
The neural networks are trained to map images to their corresponding locations within the dataset and use this mapping to estimate the locations of previously unseen images.
Notably, a landmark in the image may appear similar to multiple landmarks present on the map.
Thus, it is challenging to determine the exact landmark appearing on the image.
This fundamental limitation restricts localization accuracy and usability compared to wireless-based positioning methods. 
Moreover, the effectiveness and ability of vision-based approaches to generalize to larger and more complex environments remain key areas of ongoing research.

Another line of work in vision-based positioning is SLAM, focusing on locating and tracking robots with a map. 
Robots are equipped with various vision sensors, such as cameras, radar, and Light Detection and Ranging (LiDAR), to gather visual information about their surroundings.
Typically, the robot begins from a known initial position and continuously tracks its path while simultaneously building or updating a map of its surroundings.
This line of work has been extensively studied and has several well-regarded frameworks, such as ORB-SLAM~\cite{orb-slam}, LSD-SLAM~\cite{lsd-slam}, and DTAM~\cite{dtam}.
Generally, a SLAM framework includes Visual Odometry (VO), back-end optimization, loop closure, and mapping~\cite{chen2022overview}.
VO extracts motion-invariant features from consecutive images using techniques such as feature matching~~\cite{talukder2003real}, feature tracking~\cite{dornhege2006visual}, and optical flow~\cite{zhang2009visual}.
These features are often hand-crafted based on object-level contents~\cite{civera2011towards, salas2013slam++}, geometric shapes~\cite{engel2014lsd, newcombe2011dtam}, or visual vocabulary~\cite{sattler2011fast}. 
The back-end optimization process refines the discrete-time measurements obtained from VO and generates a globally consistent trajectory on the map. 
This process typically improves accuracy with extended Kalman filters~\cite{huang2007convergence} or particle filters~\cite{valencia2018mapping}.
However, the drift in location estimation is unavoidable and accumulates over time due to reliance on previous position estimations.
{\em Loop closure} is a technique used to detect and correct this accumulated drift. 
It identifies when the robot revisits the same location by comparing images for similarity using geometric checks~\cite{ulrich2000appearance} or bag-of-words approaches~\cite{galvez2012bags}.

% Math work on Visually based positioning 
While vision-based localization has achieved significant advances from the algorithmic perspective, its mathematical foundations have not been explored as thoroughly as those for wireless-based positioning.
In~\cite{censi2009achievable}, Fisher information and Cramer-Rao lower bounds for pose estimation are derived to address the challenges posed by the infinite-dimensional and unknown nature of the map within the context of relative sensor readings.
In contrast, \cite{rohde2016localization} considers a stochastic environment where the landmark locations form a Poisson point process (PPP).
They assume that at least three landmarks are visible and use Kalman filtering to estimate the target's location and orientation.
From this stochastic framework, an upper bound on the uncertainty in localization are derived.
In previous works, landmarks are uniquely identifiable, allowing for straightforward estimation of the target's location relative to them.
However, our setting poses a different challenge: the landmarks in our work are indistinguishable and appear in multiple locations. 
The mathematical analysis of this setting has not received much attention, other than the conference version of this work~\cite{conf1}, as well as a related work~\cite{J2}, which studied a specific localizability problem for a setup with single type of landmarks without considering any specific algorithms. The current paper provides the most comprehensive analytical treatment of this problem considering multiple types of landmarks as well as an actual algorithm that is based on geometric constraints obtained from the range measurements. More details on our technical contributions are provided in the following section.

% \chb{In~\cite{conf2}, we first rigorously pose the localization problem with landmarks having similar appearances. 
% We model the locations of these landmarks as a PPP and derive the probability of obtaining similar range measurements at candidate locations across the map. 
% This idea is extended in~\cite{J2} to include more scenarios for unmanned aerial vehicles that take aerial images and can measure the relative locations of landmarks.
% The model in \cite{conf1} has been adapted to include various types of landmarks within a finite area of interest, and the algorithm is proposed for identifying combinations of observed landmarks utilizing noisy-free range measurements.
% This paper extends the model by incorporating noisy range measurements to increase realism. 
% The landmark combinations are identified based on geometric constraints. 
% We analytically derive the probability of correctly identifying the observed landmark combinations to evaluate the performance. 
% Our contributions are outlined in the following sections.}

\subsection{Contributions}
We propose a general framework for vision-based positioning that handles landmarks with similar appearances.
The landmarks are represented by their locations and types, with those sharing similar appearances categorized under the same type.
This representation of landmarks makes it easier to store and process data for large maps.
Instead of considering each placement of landmarks on the map as a separate scenario, we treat them as instances of an underlying spatial stochastic process, as has been done with remarkable success in many other areas of research, such as wireless cellular networks~\cite{stochasticbook, cellurstochasticbook}. 
We use tools from stochastic geometry and model the locations and types of landmarks as a marked PPP. 
To make the model more realistic, we incorporate the visibility model, where landmarks are considered {\em visible} to the target if they are within a certain distance.
Instead of identifying these landmarks individually, we propose an algorithm for identifying the landmark combination observed at the target.
We prove that with three noise-free range measurements, the observed landmark combination can be almost surely identified.
Further, we develop an algorithm based on geometric constraints to identify the observed landmark combination using noisy range measurements.
Unlike existing data-driven approaches, our method does not require a training process. 
Instead, it relies on the geometric patterns formed by the landmarks compared to patterns formed by landmarks in an existing database.
We define the target as localizable if the observed landmark combination is correctly identified.
To evaluate the localization performance, we propose the localizability probability to quantify the likelihood of correctly identifying the observed landmark combination.
%Under the stochastic setting, the localizability probability denotes the likelihood of identifying the landmarks observed from an arbitrary location on the map. 
We analytically derive the expression for the localizability probability as an expectation over the joint distribution of measurements on the stochastic map.
This joint distribution of measurements depends on the spatial model of landmark locations.
These results provide benchmarks for localization in vision-based positioning. 
In addition to localization, several results developed in this paper are broadly applicable and could be applied to point processes.
Overall, our work provides a comprehensive mathematical framework and analytical results for vision-based positioning, offering insights into identifying landmarks in scenarios where the landmarks are not uniquely identifiable. 

\section{Model and Problem Formulation}\label{sec:sys-model}
This section presents the system model based on stochastic geometry, defines the concepts, and formulates the localizability problem. 
\subsection{Map of Landmarks}
Imagine an outdoor environment with various landmarks scattered throughout, such as trees, poles, and lampposts. In the real world, these landmarks are typically marked on a map and can be captured using a vision sensor, such as a camera or LiDAR. 
Each landmark observed in the visual data corresponds to a specific location on the map.
However, some landmarks, such as identical trees or lamppost, may appear visually similar and cannot be uniquely identified based solely on their appearance in the visual data. To account for this ambiguity, we treat these visually indistinguishable landmarks as having the same mark.
Mathematically, we model landmarks as marked points $\{\nbx_i, m_i\}$, where $\nbx_i \in \nbbR^2$ is the landmark location and $m_i$ is its mark in the mark space $\nbbM = \{1, \dots, M\}$. 
The random spatial patterns of these marked points are characterized by the random set $\Phi = \left\{(\nbx_i, m_i), i \in \nbbN\right\}$ on the product space $\nbbR^2 \times \nbbM$. 
In vision-based positioning, the distribution of landmarks in some scenarios follows a PPP, as shown in~\cite{rohde2016localization}. 
Therefore, it is reasonable to model the landmarks as a marked PPP, denoted by $\Phi$.
Since two points are almost surely not located at the same position in a PPP, we represent the identity of each landmark by its location.

\subsection{Visibility Regions}
Landmarks have different sizes and shapes, and their visibility distances are also different.
We assume that landmarks with the same mark are similar in size and shape. Thus, the {\em maximum visibility distance} of landmarks with the same mark $m$ can be represented as $d_m$. 
We define the landmark as being {\em visible} to the target if the distance between them is less than the maximum visibility distance.
The set of all the visible landmarks to the target located at $\nbx_0$ is 
\begin{align}
\Phi_{\rm v}(\nbx_0) = \cup_{m=1}^M \left[\Phi_{[m]} \cap \nbb\!\left( \nbx_0, d_m\right) \right],
\end{align}
where $\Phi_{[m]} = \left\{\nbx_i: (\nbx_i, m)\right\}$ is an unmarked point process that includes the locations of landmarks with mark $m$, $\nbb\!\left( \nbx_0, d_m\right)$ is a ball with radius $d_m$ centered at $\nbx_0$.
The number of visible landmarks at location $\nbx_0$ is denoted as $N = \#\{\Phi_{\rm v}(\nbx_0)\}$, which is the size of the random set $\Phi_{\rm v}(\nbx_0)$.

\subsection{Range Measurements} 
We assume that the ranges from the target to the visible landmarks can be obtained from the vision data.
The specific procedure to obtain these range measurements is not critical to our model.
For example, one can use stereo or depth cameras or even monocular vision~\cite{eigen2014depth} to obtain the range measurements.
The range measurements are denoted as a vector $\nbr = [r_1, \dots, r_N]$, where $r_i = d_i + n_i$ is the range measurement to the $i$-th landmark, $n_i$ is the measurement error, $d_i = |\nbx_i - \nbx_0|$ is the noise-free distance, $\nbx_i \in \Phi_{\rm v}(\nbx_0)$ is the location of the $i$-th landmark.
The measurement error $n_i$ is modeled as Gaussian additive noise with zero mean and variance $\sigma_i^2$.
The noise variance is assumed to be independent of noise-free distance $d_i$ and is determined by the size and shape of the landmark.
Thus, the variance depends on the mark of landmarks, denoted as $\sigma_{i}^2 = \sigma_{m_i}^2$. 
Since landmarks with the same mark are visually indistinguishable, we do not know the exact landmark with which the range measurement $r_i$ is associated.
This measurement could potentially be obtained from any visible landmark with the mark $m_i$.
In this case, $(r_i, \nbx_i)$ remains unknown, and the range measurements are associated with the marks, represented as a set of tuples
\begin{align}
    \ncalI = \left\{(r_i, m_i): r_i = |\nbx_i - \nbx_0| + n_i, \nbx_i \in \Phi_{\rm v}(\nbx_0)\right\},
\end{align}
where $m_i$ is the mark that the range measurement $r_i$ is associated with.

While the model above might seem counter-intuitive to those unfamiliar with stochastic geometry literature, the approach of treating a deterministic set of points as a realization of an appropriately chosen point process has been successfully applied across various disciplines.
The most relevant to this discussion is the study of wireless cellular networks in which one of the popular approaches is to treat the locations of wireless base stations as a realization of a Poisson point process (even though, in reality, they are an outcome of a highly complex optimization problem)~\cite{cellurstochasticbook}. 
The advantage of endowing distributions on deterministic point patterns is that it allows us to use tools from probability, specifically stochastic geometry in our case, to rigorously analyze the network-wide performance statistically.

\subsection{Problem Formulation}

\begin{figure}[t]
	\centering
	\includegraphics[width=0.32\textwidth]{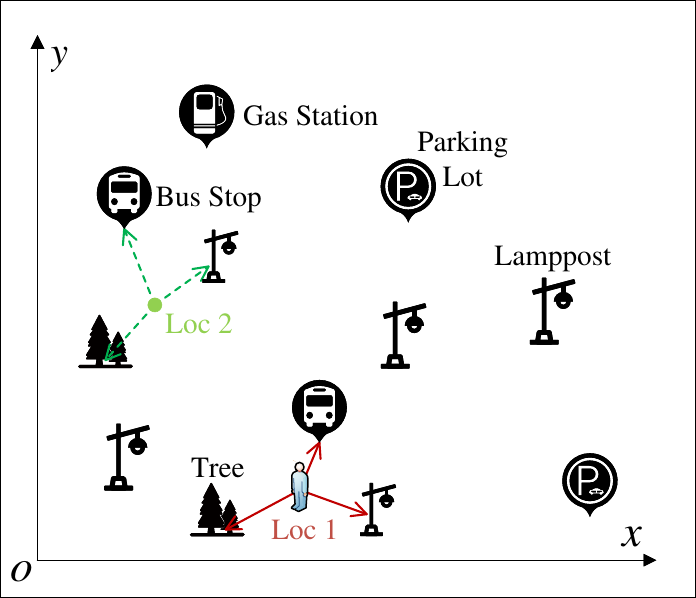}
	\caption{The illustration of the map shows five types of landmarks: lampposts, bus stops, parking lots, gas stations, and ATMs.
    Three landmarks are visible at Loc 1, but their unique locations remain uncertain because the same combination of landmarks is also visible from a different location, such as Loc 2. Although the landmarks appear identical, differences in distances, angles, or a combination of both can distinguish them.}
	\label{fig1:sysModel}
\end{figure}

To understand the technical challenges and the fundamental differences between vision-based positioning and wireless-based positioning, we illustrate an example in~\figref{fig1:sysModel}. 
The target is placed at the first location (Loc 1) with three visible landmarks. 
Since landmarks of the same type can also be observed from another location (Loc 2), it is challenging to determine the target's exact position.
This work aims to identify the observed landmark combination and estimate the target's location using measurements $\ncalI$.
We define a circular Area of Interest (AOI) $A = \nbb(\nbzero,d_a)$ with radius $d_a$, centered at the origin.
The AOI provides prior knowledge about the target's location, serving as a geographical boundary for localization.
The size of the AOI can vary significantly, ranging from tens of meters to several kilometers. 
Such prior information can be derived from services provided by base stations, Low Earth Orbit (LEO) satellites, and the Global Positioning System (GPS) infrastructure, as discussed in previous works~\cite{brosh2019accurate, zamir2010accurate}.
Given the map of landmarks within AOI, denoted as $\Phi \cap A$, and the obtained measurements $\ncalI = \left\{(r_i, m_i)\right\}$, our objective is to identify the landmark combination $c^* = [\nbx_1^*, \dots, \nbx_N^*]$ from which the range measurements $\nbr = [r_1, \dots, r_N]$ were obtained.
Once the landmark combination is correctly identified, the target's location can be estimated using the range measurements to these landmarks.
We represent all possible matching between range measurements and landmarks as the {\em combination set}, defined as
\begin{align} \label{comb-set}
	\ncalC = \left\{\left[\nbx_{1}, \dots, \nbx_{N}\right]: \nbx_{1} \in \ncalB_{m_1}, \dots, \nbx_{N} \in \ncalB_{m_N} \right\},
\end{align}
where $\left[\nbx_{1}, \dots, \nbx_{N}\right]$ is a landmark combination, $\nbx_i$ is the landmark corresponding to the range measurement $r_i$, $\ncalB_{m_i} = \Phi_{[m_i]} \cap A$ denotes all the landmarks with mark $m_i$ in AOI.
We derive the geometric constraints between visible landmarks utilizing the measurements $\ncalI$. 
The landmark combination satisfying the proposed geometric constraints is collected in the \emph{solution set} $\ncalS \subseteq \ncalC$.
The solution set $\ncalS$ may contain more than one element since the measurements are noisy, and the information provided by measurements $\ncalI$ may be insufficient to find the correct landmark combination $c^*$. 
In such cases, we uniformly at random select the estimation $\hat{c}$ from $\ncalS$.
The localization performance is evaluated by the \emph{localizability probability}, defined as the probability of correctly identifying the landmark combination utilizing $N$ measurements
\begin{align}
    \nbbP\!\left[ \hat{c} = c^* \mid N \right] = \nbbE_{\Phi} \!\left\{ \nbbP\!\left[ \hat{c} = c^* \mid N, \Phi \cap A\right] \right\},
\end{align}
which is evaluated over $\Phi$. We take advantage of this stochastic setting to evaluate localization performance across all possible geometric arrangements of landmarks.

\section{The Geometric Constraints} \label{sec:geo-feature}
In this section, we first consider the geometric constraints based on noise-free measurements. 
Then, we will extend our approach to noisy measurements. 
\begin{definition}[Minkowski Sum]
The Minkowski sum of two compact sets $\ncalA, \ncalB \in \nbbR^2$ is formed by
\begin{align}
    \ncalA \oplus \ncalB  = \left\{ \nbx + \nby \in \nbbR^2 \mid \nbx \in \ncalA, \nby \in \ncalB\right\}.
\end{align}
\end{definition}

\subsection{One Measurement}
We start with the simplest scenario involving a single measurement $\ncalI = \{(m_1, d_1)\}$. 
This measurement indicates that the target is located at a distance of $d_1$ from a landmark with mark $m_1$ within AOI.
The potential locations of the target are the union of circular regions with radius $r_1$, centered at the locations of all landmarks with mark $m_1$. 
Mathematically, this can be represented as:
\begin{align}
    \ncalB_{m_1} \oplus \partial \nbb(\nbzero, d_1),
\end{align}
where $\ncalB_{m_1}$ is the set of the landmarks in AOI with mark $m_1$, $\partial \nbb(\nbzero, d_1) = \{\nbx \mid |\nbx| = d_1\}$ is the circumference with radius $r_1$ centered at the origin. 
Without additional information, it is impossible to uniquely identify the correct landmark using only a single measurement. This is because the observed landmark cannot be differentiated from other landmarks with the same mark $m_1$.

\begin{figure}[t]
	\centering
	\includegraphics[width=0.30\textwidth]{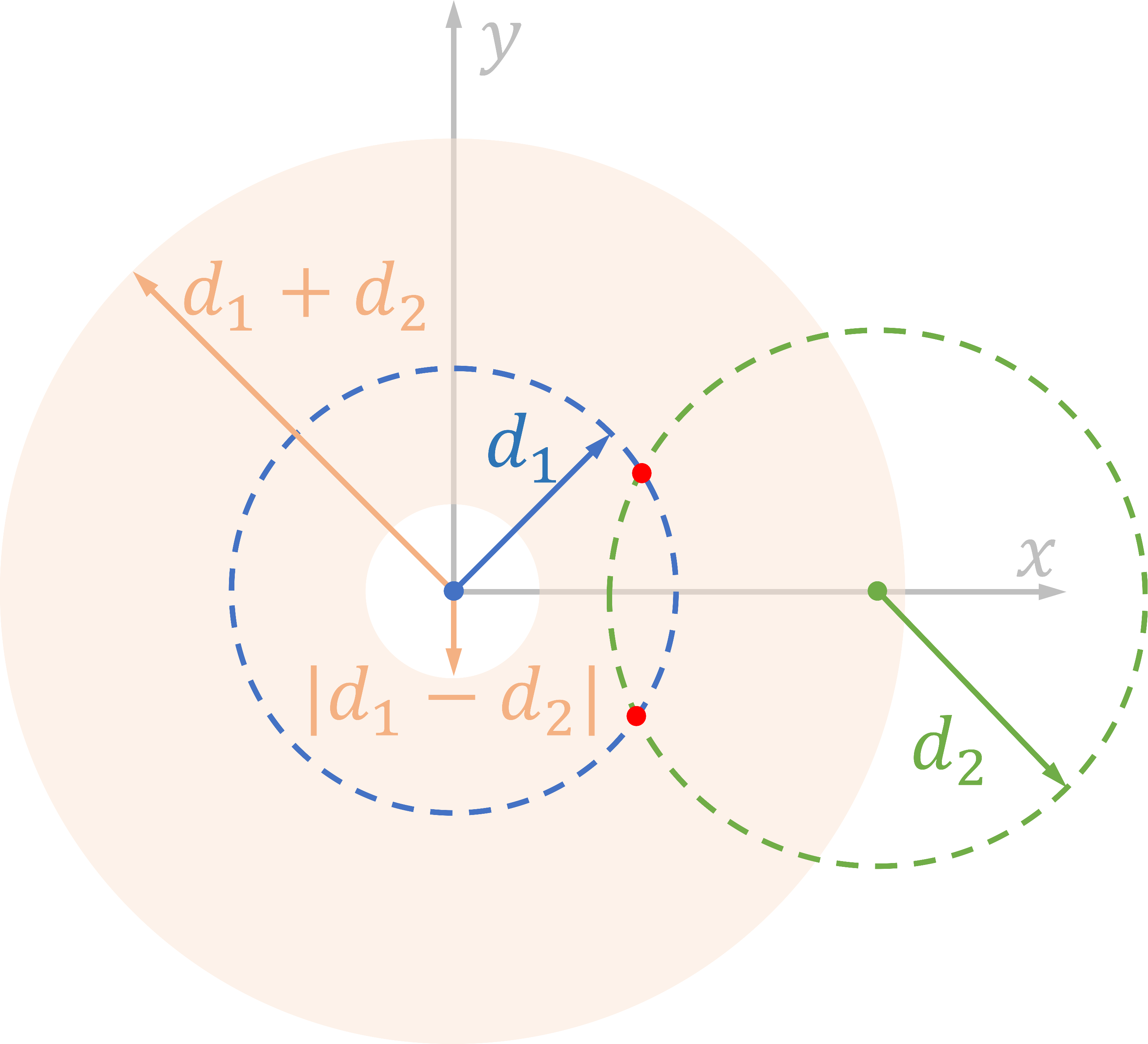}
	\caption{The illustration of possible locations for the second landmark. 
    The first landmark is fixed at the origin, represented as a blue point, and the second landmark is depicted as a green point. The target's location, marked as red points, is at a distance of $d_1$ from the first landmark and $d_2$ from the second landmark. If a target location satisfying these distance constraints exists, the possible locations of the second landmark are restricted to the orange annulus.}
	\label{fig:cir2}
\end{figure}

\subsection{Two Measurements}
Next, we consider the scenario where the target obtains two measurements $\ncalI = \{(m_1, d_1),  (m_2, d_2)\}$ from two landmarks with marks $m_1$ and $m_2$.
The geometric relationship between the target and the landmarks is illustrated in~\figref{fig:cir2}.
Without loss of generality, we place the landmark with mark $m_1$ at the origin.
The potential target locations lie on the circumference of the circle centered at the origin, represented as $\partial \nbb(\nbzero, d_1)$.
Since the distance from the target to the second landmark is $d_2$, the landmark with mark $m_2$ must be located on
\begin{align}
    \partial \nbb(\nbzero, d_1) \oplus \partial \nbb(\nbzero, d_2) = \nba(\nbzero, |d_1 - d_2|, d_1 + d_2),
\end{align}
where $\nba(\nbzero, |d_1 - d_2|, d_1 + d_2)$ represents the annulus with radii $|d_1 - d_2|$ and $d_1 + d_2$.
This result shows that the distance between two observed landmarks is constrained by $|d_1 - d_2|$ and $d_1 + d_2$.
This constraint is consistent with the geometric principles outlined in the triangle inequality. If we consider the target and the two landmarks as the vertices of a triangle, applying the triangle inequality to the triangle's edges establishes the following constraints
\begin{equation} \label{eq:tri-inequ-truth}
	|d_1 - d_2| \leq d_{12} \leq d_1 + d_2,
\end{equation}
where $d_{12} = |\nbx_1 - \nbx_2|$ is the distance between two landmarks, $d_1$ and $d_2$ are distances from the target to two landmarks.
The potential locations for the second landmark are illustrated as the orange annulus in~\figref{fig:cir2}. Since this annulus region has a nonzero Lebesgue measure, when the landmark locations follow a PPP, the probability of finding a landmark with mark $m_2$ within the annulus is nonzero. 
This suggests that the second landmark likely exists within the defined annular region, allowing the exact range measurements $d_1$ and $d_2$ to be obtained. Thus, other landmark combinations within the AOI may also result in the same measurement $\ncalI$. It is not guaranteed to uniquely pinpoint the observed landmark combination based solely on the measurement $\ncalI$.

\subsection{Three Measurements}
We then extend the scenario to include more measurements, considering the case where the target obtains range measurements from three landmarks. In this setup, we aim to demonstrate that it is feasible to determine the observed landmark combination and estimate the target location using three error-free measurements.
To support this claim, we present the following lemma, which establishes that the range measurements $d_1$, $d_2$, and $d_3$ will almost surely not correspond to any other landmark combinations.

\begin{lemma} \label{lem1}
Suppose $d_1$, $d_2$, and $d_3$ are noise-free range measurements to landmarks with marks $m_1$, $m_2$, and $m_3$, respectively. 
If the landmarks form a marked PPP on $\nbbR^2$, then it is almost surely that the range vector $\nbd = \left[d_1,d_2,d_3\right]$ can only be obtained from the observed landmark combination $c^*$. 
\end{lemma}

\begin{IEEEproof}
The geometric arrangement of the three landmarks and the target is illustrated in~\figref{fig:lem1}. 
\begin{figure}
    \centering
    \includegraphics[width=0.35\textwidth]{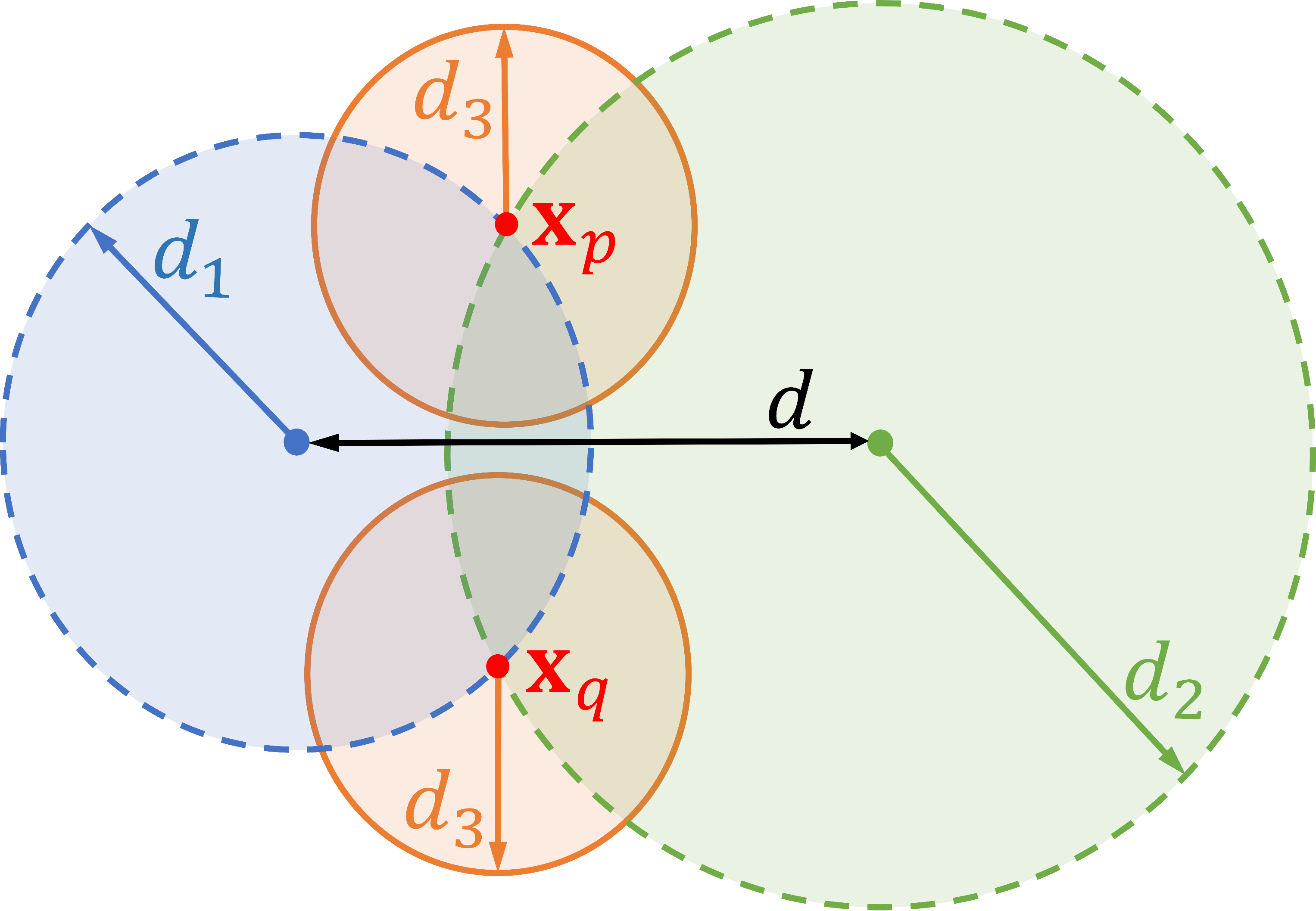}
    \caption{The illustration of possible locations for the third landmark. The blue point represents the first landmark, and the green point represents the second landmark. The red points indicate the locations of the target. The orange annulus shows the possible locations of the third landmark.}
    \label{fig:lem1}
\end{figure}
Without loss of generality, we place the first landmark at the origin. 
The second landmark is located at a distance $d$ from the first landmark, where the range $d$ satisfies the constraint $|d_1 - d_2| \leq d \leq d_1 + d_2$, by the triangle inequality.
The target obtains the range measurements $d_1$ and $d_2$ to the first and second landmarks, respectively.
Given these measurements, the target's location is constrained to lie at either of the two intersection points $\nbx_p$ and $\nbx_q$ formed by the circles centered at the two landmarks with radii $d_1$ and $d_2$, respectively.
Since the target's location is at a distance $d_3$ from the third landmark, the potential locations for the third landmark are represented by the two orange circumstances in~\figref{fig:lem1}, mathematically expressed as
\begin{align}
    \ncalX_3 = \left\{\nbx_p, \nbx_q\right\} \oplus \nbb(\nbzero, d_3).
\end{align}
Since the Lebesgue measure of $\ncalX_3$ is $\lambda(\ncalX_3) = |\ncalX_3| = 0$, when the landmark locations form a PPP, the probability that the third landmark will lie exactly on $\ncalX_3$ is almost surely zero.
Consequently, in this scenario, the probability that any other combination of three landmarks, excluding the observed combination, can obtain the same measurements $d_1$, $d_2$, and $d_3$ is almost surely zero. 
This concludes the proof.
\end{IEEEproof}

Lemma~\ref{lem1} demonstrates that the three error-free measurements obtained from the correct landmark combination are unique.
The target can be localized by finding the correct combination with unique measurements on the map. 
This result is under two assumptions: (1) the measurements are noise-free, and (2) the landmark locations form a PPP in continuous space.
In the rest of this section, we will discuss noisy measurements and investigate their pairwise geometric constraints.

\subsection{The Triangle Inequality with Noise Measurements} \label{sec:pt-pf}
Now, we consider the scenario that range measurement $r_i$ is affected by the Gaussian additive noise, represented as $r_i = d_i + n_i$, where $d_i$ is the true distance and $n_i$ is the noise.
We substitute the true distances $d_i = r_i - n_i$ into~\eqref{eq:tri-inequ-truth} and obtain the following inequalities
\begin{align} \label{eq:tri-inequ-noise}
	\begin{cases}
		r_1 + r_2 - d_{12} \geq w \\
		r_1 - r_2 + d_{12} \geq v \geq r_1 - r_2 - d_{12}
	\end{cases},
\end{align}
where $w = n_1 + n_2$ and $v = n_1 - n_2$ are the linear combinations of noise, $d_{12}$ is the distance between two landmarks. 
The value of $w$ and $v$ is unknown, but we do know the statistical properties of the random variables $W = N_1 + N_2$ and $V = N_1 - N_2$, which are linear combinations of two independent Gaussian random variables.
When the range measurements are noisy, we evaluate whether the inequalities in~\eqref{eq:tri-inequ-noise} are satisfied by a landmark pair $c_{ij} = [\nbx_i, \nbx_j]$
using the following probability, given as
\begin{align}
	\begin{split}
	A_{ij} = \big\{&\nb1\!\left(r_1 + r_2 - d_{ij} \geq W\right) \cdot \\
	&\nb1\! \left(r_1 - r_2 + d_{ij} \geq V \geq r_1 - r_2 - d_{ij} \right) = 1 \big\},
	\end{split}
\end{align}
where $\nb1(\cdot)$ is the indicator function, which takes the value $1$ if the statement inside is true and 0 otherwise.
The probability of event $A_{ij}$ is the expectation of the product of indicator functions over $W$ and $V$, written as
\begin{align} \label{eq:prob-tri-ineq}
	\nbbP\!\left[A_{ij}\right] &= \,\nbbE_{W,V}\! \big\{ 
     \mathbbm{1}\! \left(r_1 + r_2 - d_{ij} \geq W\right) \cdot  \notag\\
    & \quad \quad \quad \quad \, \mathbbm{1}\! \left(r_1 - r_2 + d_{ij} \geq V \geq r_1 - r_2 - d_{ij} \right) \big\} \notag \\
	&= \int_{-\infty}^{r_1 + r_2 - d_{ij}} \int_{r_1 - r_2 - d_{ij}}^{r_1 - r_2 + d_{ij}} f_{W,V}(w,v) \, \nrmd w \nrmd v \\
	&\begin{aligned}
    \, \, = \, \, &F_{W,V}\!\left(r_1 + r_2 - d_{ij}, r_1 + r_2 + d_{ij}\right) - \\
    &F_{W,V}\! \left(r_1 + r_2 - d_{ij}, r_1 - r_2 - d_{ij}\right),
	\end{aligned} 
\end{align}
where $f_{W,V}$ is the joint probability density function (PDF) of $W$ and $V$ and $F_{W,V}$ is the cumulative distribution function (CDF) of $W$ and $V$.
The probability of event $A_{ij}$ represents the likelihood that the triangle inequality, using measurements $r_1$ and $r_2$, holds for the landmarks located at $\nbx_i$ and $\nbx_j$, with the true distance between them being $d_{ij}$.
This probability determines whether range measurements can be obtained from a landmark combination $c_{ij} = [\nbx_i, \nbx_j]$. 
We formally define that the landmark pair $c_{ij}$ {\em satisfies} the triangle inequality in the following definition.
\begin{definition}
    If the triangle inequalities in~\eqref{eq:tri-inequ-noise} are satisfied by a landmark pair $c_{ij}$ with high probability, denoted as $\nbbP[A_{ij}] \geq T$, the landmark combination $c_{ij}$ satisfy the geometric constraint using noisy range measurements $r_1$ and $r_2$. 
\end{definition}

\section{The Localizability Probability with Two Measurements} \label{sec:two-ranges}
This section uses the proposed geometric constraints to estimate the observed landmark combination.   
First, we examine the landmark combinations in $\ncalC$ and add those that satisfy the proposed geometric constraints to the solution set.
The solution set $\ncalS$ is represented as
\begin{equation} \label{eq:solution-set}
	\ncalS = \left\{ c_{ij} \mid c_{ij} \in \ncalC, P[A_{ij}] > T\right\},
\end{equation}
where the landmark combination $c_{ij}$ contains two landmarks located at $\nbx_i$ and $\nbx_i$, the event $A_{ij}$ represents the that the triangle inequalities are satisfied by $c_{ij}$.
To evaluate the landmark combinations added to the solution set, we are interested in the following two probabilities:

\subsection{True Positive Rate (TPR)}
The {\em true positive rate} $p_t$ represents the probability that the correct combination $c^*$ is included in the solution set, defined as
\begin{equation} \label{eq:TPR}
    p_t =  \nbbP[I_{D^*} = 1 \mid \nbr, \nbm ] = \nbbE_{ D^* \mid \nbr, \nbm } \left\{\nb1\!\left(\nbbP[A^*] \geq T\right) \right\},
\end{equation}
where the random variable $D^*$ represents the distance between landmarks in $c^*$, the binary variable $I_{D^*} = \nb1\!\left(\nbbP[A^*] \geq T\right) = 1$ indicates whether the correct combination $c^*$ satisfies the triangle inequalities, $\nbr = [r_1, r_2]$ are the range measurements, $\nbm = [r_1, r_2]$ are the marks corresponding to the range measurements.
We use the stochastic setting to evaluate the true positive rate over all possible landmark placements. 
Thus, the true positive rate is an expectation over the random variable $D^*$.
This result can be interpreted as the average rate at which the correct combination $c^*$ {\em satisfies} the triangle inequality based on range measurements $r_1$ and $r_2$.

\subsection{False Positive Rate (FPR)}
Another important probability is the {\em false positive rate}, which represents the likelihood that any landmark combination $c \in \ncalC$ is incorrectly included in the solution set. It is defined as
\begin{equation} \label{eq:fpr}
    p_f =  \nbbP [I_{D} = 1 \mid \nbr, \nbm] = \nbbE_{ D \mid \nbr, \nbm } \left\{\nb1\!\left(\nbbP[A] \geq T\right)\right\},
\end{equation}
where the $D$ represents the distance between the landmarks in the combination, $I_{D} = \nb1\!\left(\nbbP[A] \geq T\right) = 1$ represents the landmark combination satisfies the triangle inequalities.
We interpret $p_f$ as the average rate at which landmark combinations within AOI {\em satisfy} the proposed triangle inequality based on range measurements $r_1$ and $r_2$.
We aim to minimize the false positive rate $p_f$ to ensure that incorrect landmark combinations are excluded from the solution set $\ncalS$ as much as possible.

Both TPR and FPR are the expectations over the distances between landmarks, which depend on the placement of the landmarks. 
Higher measurement noise decreases $p_t$, thereby decreasing accuracy, and increases $p_f$, making it more likely for incorrect landmark combinations to be included. These metrics are critical for deriving the {\em localizability probability}, which quantifies the likelihood of successfully determining the target's location.

Under the assumption that the landmarks form a homogeneous marked PPP, we further derive the analytic expression for $p_t$ and $p_f$. 
Before that, we first give the distribution of the distance in the following lemma.
\begin{lemma} \label{lem2}
Given the number of landmarks within AOI, the distribution of the distance between two arbitrary landmarks $c_{ij} = \left\{\nbx_i, \nbx_j\right\}$ is
\begin{equation} \label{eq:lem2}
    f_D(s) = \begin{cases}
            \frac{2d_a^2 s \left(\pi - s \sin^{-1}\left(\frac{s}{2d_a}\right) \right) - s^2 \sqrt{4d_a^2 - s^2}}{\pi d_a^4} \quad 0\leq s\leq 2d_a,\\ 
            0 \quad {\rm others}.
    \end{cases} \notag
\end{equation}
\end{lemma} 
\begin{IEEEproof}
For a PPP, when conditioned on the number of points in a bounded region, these points are independently and uniformly distributed~\cite{stochasticbook}. 
As a result, calculating the distance distribution between two points reduces to determining the distribution of the distances between two arbitrary points within the AOI.
This is a well-established result in stochastic geometry and is thoroughly discussed in~\cite{distanceDistribution}.
\end{IEEEproof}

Next, we derive the distance distribution between the correct landmark combinations. 
Since the number of visible landmarks may not always be two, this distribution depends on the specific method used to select the subset of visible landmarks from which the measurements are obtained.
The methods for selecting this subset are called the {\em observation policy}. In this work, we consider two observation policies: the random policy and the nearest policy. However, our analytical approach is not restricted to these two policies. Various observation policies could be employed, and the corresponding distributions can be derived accordingly.
The random policy refers to obtaining measurements from arbitrary landmarks within the visibility distance.
In this setting, the PDF of $D^*$ is identical to that of $D$ since the probability of selecting any pair of the observed landmarks is the same.
In contrast, the nearest policy selects the landmarks closest to the target, resulting in smaller range measurements.
Since smaller range measurements lead to a tighter triangle inequality, the nearest policy outperforms the random policy regarding localizability probability.
Next, we will derive the {\em localizability probability} based on the TPR and the FPR.
We will compare the above two observation policies under the same parameter setting by evaluating the TPR $p_t$, the FPR $p_f$, and the localizability probability in the numerical result section.

\subsection{The Localizability Probability} \label{loc-prob}
To localize the target, we identify the landmark combination from which the measurements are obtained. 
The target is considered {\em localizable} if the correct landmark combination is successfully identified.
The observed landmark combination is estimated using the geometric constraints proposed in Section~\ref{sec:geo-feature}.
To evaluate the localization performance analytically, we derive the probability of correctly estimating the landmark combination based on the geometric constraints. This probability is referred to as the {\em localizability probability}.
Since the measurements are noisy, the number of solutions in $\ncalS$ might not necessarily be one. Thus, we select the combination estimation using the following approach:

Since the measurements are noisy, the number of solutions in $\ncalS$ might not be exactly one. Therefore, we select the combination estimate using the following approach:
When the solution set size $|\ncalS|$ is 1,  the estimated landmark combination is the only one in the set.
If there is more than one solution in the solution set, we uniformly at random select one solution from $\ncalS$.
With this approach, the size of the solution set $\ncalS$ affects the localization performance. 
Thus, to derive the localizability probability, we first construct the distribution of the number of elements in $\ncalS$, as presented in the following lemma:

\begin{lemma} \label{lem3}
Given the measurements $\ncalI = \left\{\nbr, \nbm\right\}$, the number of potential landmark combinations $|\ncalC|$ and whether the correct combination is in $\ncalS$, the conditional probability mass function (PMF) of $|\ncalS|$ is 
\begin{align} \label{eq:lem3}
    \nbbP&\!\left[ |\ncalS| = k \mid \nbr, \nbm, |\ncalC| = m, I^*\right] \notag \\
    &=\begin{cases}
        \binom{m-1}{k} p_f^{k} (1 - p_f)^{m - k - 1}, \quad & I^* =0,\\
        \binom{m-1}{k-1} p_f^{k-1} (1 - p_f)^{m - k}, \quad & I^* =1,
    \end{cases}
\end{align}
where $\nbr = \left[r_1, r_2\right]$ are the range measurements, $\nbm = \left[m_1, m_2\right]$ are the corresponding marks of the range measurements, $p_f$ is the FPR defined in~\eqref{eq:fpr}, and $I^*$ is the indicator representing whether the correct landmark combination is included in the solution set.
\end{lemma} 

\begin{IEEEproof}
The size of the solution set $\ncalS$ is the number of landmark combinations in $\ncalC$ that satisfy the proposed geometric constraints.

We use a binary indicator $I_{ij}$ to represent whether the landmark combination $c_{ij} = \left\{\nbx_i, \nbx_j\right\} $ satisfies the geometric constraints, defined as:
\begin{align}
    I_{ij} = \mathbbm{1}\!\{\nbbP\left[A_{ij}\right] \geq T\}.
\end{align}
Assuming the distances between landmarks in the combinations are independent and identically distributed (i.i.d.), their corresponding indicators $I_{ij}$ are i.i.d. Bernoulli random variables. The number of elements in $\ncalS$ is the sum
\begin{align}
    |\ncalS| = \sum_{c_{ij} \in \ncalC} I_{ij},
\end{align}
which is a binomial random variable. 
The conditional probability of $|\ncalS|$ is 
\begin{align} \label{eq:solution-size}
    \nbbP\!&\left[|\ncalS| = k \mid \nbr, \nbm, |\ncalC| = m, I^* \right] \notag\\
    &= \nbbP\!\left[\sum_{c_{ij} \in \ncalC} I_{ij} = k \,\Bigg|\, \nbr, \nbm,  |\ncalC| = m, I^*\right],
\end{align}
where $I^*$ indicates whether the correct combination $c^*$ satisfies the geometric constraints and is contained in the solution set.
When $I^* = 0$, meaning $c^*$ is not included in the solution set, the above equation simplifies to
\begin{align}
    \nbbP&\!\left[|\ncalS| = k \mid \nbr, \nbm, |\ncalC| = m, I^* = 0\right] \notag \\
    &= \nbbP\left[\sum_{c_{ij} \in \ncalC \backslash \left\{c^*\right\}} \! I_{ij} = k \mid \nbr, \nbm,  |\ncalC| = m\right]\\
    &\,\begin{aligned}
    \overset{(a)}{=} &\binom{m-1}{k} \bigg\{\nbbP \! \left[I_{ij} = 1 \mid \nbr, \nbm\right]\bigg\}^k \\
    &\cdot \bigg\{\nbbP\!\left[ I_{ij} = 0 \mid \nbr, \nbm\right]\bigg\}^{m - k - 1}
    \end{aligned}\\
    &\overset{(b)}{=}  \binom{m-1}{k} p_f^{k} \left(1 - p_f\right)^{m - k - 1},
\end{align}
where (a) follows from the fact that $I_{ij}$ is i.i.d., (b) follows from the definition of the FPR in~\eqref{eq:fpr}.

When $I^* = 1$, meaning that $\ncalS$ contains the correct combination $c^*$. Equation~\eqref{eq:solution-size} can be expressed in an alternative form as follows:
\begin{align}
    &P\!\left[|\ncalS| = k \mid \nbr, \nbm, |\ncalC| = m, I^* = 1\right] \notag \\
    &\overset{(a)}{=} P\!\left[\sum_{c_{ij} \in \ncalC \backslash \left\{c^*\right\}} \! I_{ij} = k-1 \mid \nbr, \nbm,  |\ncalC| = m\right] \\
    &= \binom{m-1}{k-1} p_f^{k-1} (1 - p_f)^{m - k},
\end{align}
where (a) follows from the fact that $k - 1$ landmark combinations exist in $\ncalC \backslash {c^*}$ satisfying the geometric constraints if $c^* \in \ncalS$. 
This completes the proof
\end{IEEEproof}

Lemma~\ref{lem3} provides the conditional PMF of the number of elements in the solution set.

With this result, we can derive the probability of correctly estimating the observed landmark combination. This probability, referred to as the localizability probability, is defined as the expectation over the measurements $\ncalI = \left\{ \nbR, \nbM \right\}$ and the size of combination set $|\ncalC|$, given as
\begin{equation} \label{eq:loc-prob}
    \nbbP\!\left[ \hat{c} = c^* \mid  N = 2 \right] = \mathbb{E}_{\nbR, \nbM, |\ncalC|}\! \left\{ \nbbP\!\left[ \hat{c} = c^* | \nbr, \nbm, |\ncalC| \right]\right\}.
\end{equation}
The localizability probability quantifies the average likelihood of correctly estimating the observed landmark combination across different realizations of the placements of landmarks.

Building upon the previous lemmas, we present this paper's main result in the following theorem.
\begin{theorem} \label{the1}
When utilizing the geometric constraints established by two measurements, the localizability probability is
\begin{equation} \label{eq:the1}
    \nbbP\!\left[ \hat{c} = c^* | N = 2 \right] = \mathbb{E}_{\nbR, \nbM, |\ncalC|}\! \left\{p_t \cdot \frac{1 - \left(1 - p_f\right)^{|\ncalC|}}{|\ncalC| \cdot p_f} \right\}.
\end{equation}
\end{theorem}

\begin{IEEEproof}
Using the law of total probability, Equation~\eqref{eq:the1} can be written as
\begin{align}
    \nbbP& \!\left[\hat{c} = c^* \mid \nbr, \nbm, |\ncalC| = m \right] \notag\\
    &\,\begin{aligned}
    \overset{(a)}{=} \! \sum_{I = 0}^{1}  \sum_{k=0}^{m} \! \bigg\{ &\nbbP\!\left[\hat{c}\! = \!c^* \mid \nbr, \nbm, |\ncalC|\! =\! m, |\ncalS|\! =\! k, I^*\right] \\
    & \cdot \nbbP\!\left[|\ncalS| = k \mid \nbr, \nbm, |\ncalC| = m, I^*\right] \\ 
    & \cdot \nbbP\!\left[I^* \mid \nbr, \nbm, |\ncalC| = m\right] \bigg\}
    \end{aligned}\\
    &\,\begin{aligned}
    \overset{(b)}{=} \sum_{k=0}^{m} \bigg\{&\nbbP\!\left[\hat{c}\! =\! c^* \mid \nbr, \nbm, |\ncalC|\! =\! m, |\ncalS|\! =\! k, I^*\! =\! 1\right] \\
    & \cdot \nbbP\!\left[|\ncalS| = k \mid \nbr, \nbm, |\ncalC| = m, I^* = 1\right] \\
    & \cdot \nbbP\!\left[I^* = 1 \mid \nbr, \nbm, |\ncalC| = m\right] \bigg\}
    \end{aligned} \label{eq:the1-b}\\
    & \overset{(c)}{=} \sum_{k=1}^{m} \frac{p_t}{k} \cdot \nbbP\!\left[|\ncalS| = k \mid \nbr, \nbm, |\ncalC| = m, I^* = 1\right],
\end{align}
where (a) follows from the law of total probability, (b) follows from the fact that when the correct combination is not in $\ncalS$, it is impossible to correctly estimate the observed landmark combination, (c) follows from the condition that when $c^* \in \ncalS$, both $|\ncalC|$ and $|\ncalS|$ are at least $1$. 

The first component in~\eqref{eq:the1-b} is determined by the estimation process, in which we uniformly at random select the estimation $\hat{c}$ from $\ncalS$. 
Thus, the probability of selecting the correct combination is given by
\begin{gather}
    \nbbP\! \left[\hat{c} = c^* \mid \nbr, \nbm, |\ncalC| = m, |\ncalS| = k, I^* = 1\right] \notag\\
    =\begin{cases}
        \frac{1}{k}, \quad m\geq k \geq 1 \\
        0, \quad {\rm others}
    \end{cases},
\end{gather}
where $I^* = 1$ indicates that the correct solution is in $\ncalS$.

The second component in~\eqref{eq:the1-b} has been derived in Lemma~\ref{lem3}, while the third component is provided in~\eqref{eq:TPR} as the TPR $p_t$. Now, we can combine these results to obtain the final expression for the localizability probability
\begin{equation}
    \label{eq:the1:result}
    \nbbP\!\left[ \hat{c} = c^* \mid \nbr, \nbm, |\ncalC| = m\right] = p_t \cdot \frac{1 - \left(1 - p_f\right)^m}{m p_f}.
\end{equation}
With the definition in~\eqref{eq:loc-prob}, the localizability probability can be written in the form of Equation~\eqref{eq:the1}. This completes the proof.
\end{IEEEproof}

%%%%%%%%%%%%%%%%%%%% Note Here %%%%%%%%%%%%%%%%%%%%%%%
% Note: Change the expression here. Emphasis on the fact that Theorem 1 could be used in two ways: 1. Semi-Analytical: Generate the joint distribution from simulation. 2. Analytical: Take the expectation over the joint distribution.

Theorem~\ref{the1} presents the analytical expression for the localizability probability when utilizing the geometric constraints derived from two measurements.
This result, in the context of a stochastic setting, quantifies the probability of correctly estimating the observed landmark combination across all possible landmark placements. Using the stationarity of the homogeneous PPP, this probability also corresponds to the proportion of spatial locations where a landmark can be successfully localized. As a result, it serves as a fundamental benchmark for evaluating localization performance.
The result in Theorem~\ref{the1} is expressed in an expectation form and relies on the joint distribution of $\nbR$, $\nbM$, and $|\ncalC|$. To proceed further, we need to make assumptions about the observation policy, which will primarily impact the joint distribution of the three random variables. While we can conceptually proceed for a variety of observation policies, we illustrate the whole process for the the {\em random policy} for which it is possible to derive explicit expression for the joint distribution. Analysis will proceed in the similar manner for other policies, but with varying levels of tractability. Using the derived joint distribution, we specialize the general result of Theorem~\ref{the1} to the selected policy in Theorem~\ref{the3}.

\subsection{The Joint Distribution of $\nbR, \nbM, |\ncalC|$} 
To obtain the close form of the analytical result presented in Theorem~\ref{the1}, we derive the joint distribution of measurements $\nbR, \nbM$ and the number of elements in the combination set $|\ncalC|$.
Before that, we specify how the two landmarks of interest were chosen from among the visible landmarks. 
Consider $M$ types of landmarks with visible distances $d_{1}, d_{2}, \dots, d_{M}$, respectively. The locations of the $m$-th type of landmark form a PPP $\Phi_m$ with density as $\lambda_m$, where $m \in \left\{1, 2, \dots, M\right\}$. 
We assume that the two landmarks, and consequently the two measurements, are independently and uniformly at random selected from the set of visible landmarks.
Under this setup, the joint distribution is expressed as
\begin{align} 
    \label{eq:joint-prob-3term}
    \begin{aligned}
        f_{\nbR, \nbM, |\ncalC|}&\!\left( \nbr, \nbm, n\right)  \\
        =\ &\nbbP\!\left[ \nbM = \nbm\right] \cdot f_{\nbR | \nbM}\left[ \nbr \mid \nbm\right] \cdot \nbbP[|\ncalC| = n \mid \nbm],
    \end{aligned}
\end{align}
which follows from the fact that the two landmarks from which the measurements are obtained are selected uniformly at random.
Thus, when conditioned on the landmark types, the joint distribution of range measurements is independent of the size of $\ncalC$. 
In the rest of this section, we will derive expressions for the three terms in~\eqref{eq:joint-prob-3term}. The PMF of the mark of the landmark is given in the following lemma.

\begin{lemma} \label{lem4}
    The marginal distribution of landmark types that correspond to range measurements is
    \begin{align}
        \nbbP\!\left[ \nbM = \left[p, q\right]\right] 
        = \frac{\Lambda_p \cdot \Lambda_q }{\left( \sum_{m = 1}^{M} \Lambda_m \right)^2},
        \label{eq:lem4}
    \end{align}
    where $\nbM = \left[M_1, M_2\right]$, $M_1$ and $M_2$ are random variables representing the mark of the landmarks that measurements obtained from, $p$ and $q$ are two indices of the landmark types, and $\Lambda_p = \lambda_p \pi d_p^2$.
\end{lemma} 
\begin{IEEEproof}
    Because the landmarks are selected uniformly at random from the visibility region, the two measurements (and hence their corresponding marks) are independently and identically distributed, given as
    \begin{align}
        \nbbP\!\left[ \nbM = \left[p, q\right]\right]  = \nbbP\!\left[ M_{1} = p\right] \cdot \nbbP\!\left[ M_{2} = q\right].
    \end{align}
    Now, it is sufficient to characterize the marginal distribution of one of the landmark types, for example, $\nbbP\!\left[ M_{1} = p\right]$.
    Since landmarks of different types have different visible distances, we scale the PPP $\Phi_{[1]}, \dots, \Phi_{[M]}$, such that visible landmarks of different types are mapped onto a unit circle centered at the target location $\nbx_0$, denoted as $\nbb\!\left(\nbx_0, 1\right)$. 
    The scaled (or transformed) PPP is denoted as $\tilde{\Phi} = \bigcup_{m=1}^{M} \tilde{\Phi}_{[m]}$, where $\tilde{\Phi}_{[m]}$ is the scaled PPP of the $m$-th type of landmarks with density $\lambda_m^\prime = \lambda_m d_m^2$. 
    Instead of choosing visible landmarks with different visible distances, we equivalently uniformly at random select marked points from $\tilde{\Phi} \cap \nbb\!\left(\nbx_0, 1\right)$. 
    Thus, the mark of the selected points is independent of their locations, given by
    \begin{equation}
        \label{eq2:lem4}
        \nbbP\!\left[ M_{1} = p\right]  = \frac{\Lambda_p}{\sum_{m = 1}^{M}\Lambda_m},
    \end{equation}
    where $\Lambda_p = \lambda_p \pi d_p^2$.
    The distribution of mark $M_2$ is identical. This completes the proof.
\end{IEEEproof}

Next, we give the conditional PDF of the range measurements in the following lemma.
\begin{lemma} \label{lem5}
    The conditional joint distribution of distance measurements is 
    \begin{align}
        f_{\nbR | \nbM}\!\left( \nbr \mid \nbm\right)  = \frac{2 r_1 \mathbbm{1}_{r_1} ((0, d_p])}{d_p^2} \cdot \frac{2 r_2 \mathbbm{1}_{r_2} ((0, d_q])}{d_q^2}. \label{eq:lem5}
    \end{align}
\end{lemma} 

\begin{IEEEproof}
    Since the two landmarks are selected uniformly at random, the resulting two measurements are independent. Again, considering the range measurement's marginal distribution is sufficient.
    When conditioned on the landmark type, its corresponding visibility distance is known. 
    The location of the selected landmark is uniformly distributed within the circle with radius $d_p$, which gives
    \begin{equation}
        f_{R_1 \mid M_1}(r_1 \mid  p) = \frac{2 r_1 \mathbbm{1}_{r_1} ((0, d_p])}{d_p^2},
    \end{equation}
    which is a straightforward consequence of the PPP assumption of the landmark locations. This completes the proof.
\end{IEEEproof}

Then, we provide the expression for the last term in~\eqref{eq:joint-prob-3term}.
\begin{lemma} \label{lem6}
    The conditional PMF of the size of the combination set $\ncalC$ is 
    \begin{align}
        &\nbbP\!\left[ |\ncalC| = n \mid \nbM = \left[p, q\right]\right]  \notag \\
        &\begin{aligned}
            = \sum_{n_1 n_2 = n} &\bigg\{\sum_{v = 1}^{n_1}  \frac{c_p^{n_1 - v} b_p^{v} (b_p -  a)^{-v} }{(n_1 - v)! } \\
            & \cdot \frac{\Gamma(v) - \Gamma(v, b_p-a) }{\Gamma(v)}\exp\!\left(  - a - c_p\right) \!\bigg\}\\
            &\cdot \bigg\{\sum_{u = 1}^{n_2} \frac{c_q^{n_2 - u} b_p^{u} (b_q-a)^{-u} }{(n_2 - u)! } \\
            &\cdot \frac{\Gamma(u) - \Gamma(u, b_p-a) }{\Gamma(u)}\exp\!\left( - a - c_q\right) \!\bigg\},
        \end{aligned}
    \end{align}
    where $p$ and $q$ are different landmark marks, $a = \sum_{m = 1}^{M} b_m $, $b_p =  \lambda_p \pi d_p^2 $ and $c_p = \lambda_{p} \pi( d_a^2 - d_p^2)$.
\end{lemma} 

\begin{IEEEproof}
    When two distance measurements are obtained from two landmarks of different types, the size of the combination set is 
    \begin{equation}
        |\ncalC| = N_p \cdot N_q,
    \end{equation}
    where $N_p = \#\! \left\{\ncalB_p\right\}$ are the number of landmarks of type $p$ within the AOI, $\ncalB_p = \Phi_{[p]} \cap A_\nbx$ represents the landmarks of type $p$ within AOI and $A_\nbx$ represents the AOI.
    The corresponding measurements are independent since the landmarks are selected uniformly at random. The numbers of the landmarks of type $p$ and $q$ within AOI are independent, given as
    \begin{align}
        \begin{aligned}
            &\nbbP\!\left[ N_p  = n_p, N_q  = n_q \mid M_1 = p, M_2 = q\right] =\\
            &\nbbP\!\left[  N_p = n_p \mid M_1 = p\right] \nbbP\!\left[ N_q = n_q \mid M_{2} = q\right].
        \end{aligned}
    \end{align}
    Then, we use the total probability law and write the conditional PMF of $N_p$ as  
    \begin{align}
        \label{eq:pmf-np}
        &\nbbP\!\left[ N_p  = n_p \mid M_1 = p\right] \notag\\
        &\begin{aligned}
            =\ \sum_{v = 1}^{n_p} \bigg\{&\nbbP\!\left[ N_p = n_p \mid M_1 = p,  V_p = v\right] \\
            &\cdot \nbbP\!\left[ V_p = v \mid M_1 = p \right]\bigg\},
        \end{aligned}
    \end{align}
    where $ V_p = \#\! \left\{\Phi_{[p]} \cap \nbb\!\left( \nbx_0, d_p\right) \right\}$ is the number of visible landmarks of type $p$, $\nbb\!\left( \nbx_0, d_p\right)$ is the visibility region corresponding to type $p$ landmarks.
    The first term in~\eqref{eq:pmf-np} denotes that the conditional PMF of the number of landmarks within mark $p$ in the AOI, given that 
    there are already $V_p = v$ landmarks with mark $p$ in the visibility region.
    The second term represents the PMF of the number of visible landmarks with the mark $p$.
    Now, we further write the first term in~\eqref{eq:pmf-np} as
    \begin{align}
        &\nbbP\!\left[  N_p  = n_p \mid M_1 = p,  V_p = v\right]  \notag \\
        &= \nbbP\!\left[  H_p  = n_1 - v \mid M_1 = p\right]\\
        &= \frac{c_p ^{n_1 - v}}{\left( n_1 - v\right) !} \exp\!\left(- c_p\right),
    \end{align}
    where $H_p = N_p - V_p$ is a random variable representing the number of hidden landmarks with mark $p$, which are within AOI but outside of the visibility region, and $c_p = \lambda_{p} \pi\!\left( d_a^2 - d_p^2\right)$. This result is a direct consequence of the PPP assumption~\cite{stochasticbook}.
    
    Now, we left the second term in~\eqref{eq:pmf-np}. This term represents the conditional PMF of the number of landmarks with the mark $p$ in the visibility region. To derive this probability, we first derive their joint probability. We use $W$ to represent the total number of visible landmarks, defined as
    \begin{align}
        W &= \#\! \left\{\Phi_v\!\left(\nbx_0\right)\right\} \\
        &= \#\! \left\{\cup_{m=1}^M \left[\Phi_{[m]} \cap \nbb\!\left( \nbx_0, d_m\right) \right] \right\} \\
        &= \sum_{m = 1}^{M} \#\!\left\{ \Phi_{[m]} \cap \nbb(\nbx_0,d_m)\right\} \\
        &= \sum_{i = 1}^{M} \#\! \left\{ \tilde{\Phi}_{[m]} \cap \nbb(\nbx_0,1)\right\} \\
        &= \#\! \left\{ \cup_{m=1}^M \left[\tilde{\Phi}_{[m]} \cap \nbb(\nbx_0,1) \right]\right\}\\
        &= \#\! \left\{ \tilde{\Phi} \cap \nbb(\nbx_0,1)\right\},
    \end{align}
    where $\tilde{\Phi} = \cup_{m=1}^M \tilde{\Phi}_{[m]}$ is the scaled unmarked PPP defined in the proof of Lemma~\ref{lem4}, $\tilde{\Phi}_{[m]}$ is the PPP of landmarks with mark $m$.
    Now, the joint PMF of random variables $M_1$, $V_p$ and $V$ can be written as
    \begin{align} \label{eq:lem6-proof}
        \nbbP&\!\left[ M_1 = p, V_p = v, W = w\right]  \notag\\
        &= \nbbP\!\left[ M_1 = p \mid V_p = v, W = w\right] \nbbP\!\left[ V_p = v, W = w\right],
    \end{align}
    where the first component $\nbbP\!\left[ M_1 = p \mid V_p = v, W = w\right]$ represents the probability that the selected landmark has the mark $p$, given that the landmark is uniformly at random chosen from $W$ visible landmarks, and there are $V_p$ with mark $p$. This probability is straightforward, given as
    \begin{equation}
        \nbbP\!\left[ M_1 = p \mid V_p = v, W = w\right]  =  \frac{v}{w}, \quad 0 \leq v \leq w.
    \end{equation}
    The second component in~\eqref{eq:lem6-proof} is the joint PMF of $V$ and $W$, written as
    \begin{align}
        \nbbP&\!\left[ V_p = v, W = w\right]   \notag \\
        &\overset{(a)}{=} \nbbP\!\left[ W - V = w - v \mid V_p = v\right] \nbbP\left[V_p = v\right] \notag\\
        &= \frac{(a - b_p)^{w - v}}{(w - v)!} \exp\!\left(  -a + b_p \right) \cdot \frac{b_p^{v}}{v!} \exp\!\left(- b_p\right),	
    \end{align}
    where (a) followed from the independence of the marks of different landmarks, $a = \sum_{m = 1}^{M} b_m$ and $b_p =  \lambda_p \pi d_p^2$.
    
    Now, with the derived joint distribution of $M_1$, $V_p$ and $W$,
    We take the summation and write the second component in~\eqref{eq:pmf-np}
    \begin{align}
        \nbbP&\!\left[ V_p = v \mid M_{1} = p \right]  \notag \\
        =& \!\sum_{w = v + 1}^{\infty}\! \frac{ \nbbP\!\left[ M_1 = p, V_p = v, W  = w\right] }{\nbbP \left[ M_1 = p\right] }\\
        =&\frac{a b_p^{v} \left[ \Gamma\!\left( v\right)  - \Gamma\!\left( v, b_p - a\right) \right] }{b_p (b_p-a)^{v}\Gamma(v)} \exp\!\left( -a\right).
    \end{align}
    With the expressions for two components, equation~\eqref{eq:pmf-np} is summarized as
    \begin{align}
        \nbbP&\!\left[ N_p  = n_1 \mid M_1 = p \right] \notag \\
        &\begin{aligned}
            = \sum_{v = 1}^{n_1} \bigg\{&\frac{c_p^{n_1 - v} b_p^v }{\left( b_p - a\right) ^{v} \left( n_1 - v\right) ! } \\
            &\cdot \frac{\Gamma\!\left( v\right)  - \Gamma\!\left( v, b_p-a\right)  }{\Gamma\!\left( v\right) } \exp\!\left(  -a - c_p\right) \bigg\}.
        \end{aligned}
    \end{align}
    
    Since two landmarks are selected uniformly at random, the probability 
    $\nbbP\!\left[ V_q = n_2 \mid M_2 = q\right)$ can be derived with the same method. 
    Because of the independence of $V_p$ and $Vq$, we can easily take the summation and construct the conditional PMF of $|\ncalC|$. This completes the proof.
\end{IEEEproof}

Using the result from Lemmas~\ref{lem4}, \ref{lem5}, \ref{lem6}, we obtain the joint distribution of the measurements $\nbR$, their corresponding types $\nbM$ and the size of combination set $|\ncalC|$, given next.

\begin{theorem}
    \label{the3}
    The joint distribution of $\nbR$, $\nbM$, $|\ncalC|$ is given as
    \begin{align}
        &f_{\nbR, \nbM, |\ncalC|}(\nbr, \nbm,  n)= \frac{4 r_1 r_2 \lambda_p \lambda_q \mathbbm{1}_{r_1} ((0, d_p]) \mathbbm{1}_{r_2} ((0, d_p])}{a^2} \notag\\
        &\begin{aligned}
            \cdot \sum_{n_1 n_2 = n}\Bigg\{ \bigg\{&\sum_{v = 1}^{n_1} \frac{c_p^{n_1 - v} b_p^{v} }{\left( b_p-a\right) ^{v} \left( n_1 - v\right) ! }  \\
            &\cdot \frac{\Gamma\!\left( v\right)  - \Gamma\!\left( v, b_p-a\right)  }{\Gamma\!\left( v\right) }\exp\!\left( -a - c_p\right)  \bigg\} \\
        \end{aligned}\\
        &\quad \quad \quad \quad\begin{aligned}
            \cdot \bigg\{&\sum_{u = 1}^{n_2} \frac{c_q^{n_2 - u} b_q^{u} }{\left( b_q-a\right) ^{u} \left( n_2 - u\right)! } \\
            &\cdot \frac{\Gamma\!\left( u\right)  - \Gamma\!\left( u, b_q-a\right)  }{\Gamma\!\left( u\right)}\exp\!\left( -a - c_q\right)  \bigg\}\Bigg\}. \notag
        \end{aligned}
    \end{align}
\end{theorem}
This theorem, combined with the result in Theorem~\ref{the1}, fully characterizes the localizability probability, marking one of the key technical contributions of this paper.

\section{Localizability with More Than Two Measurements} \label{sec:three-ranges} 
In the previous section, we derived the localizability probability when two range measurements are available. 
Additionally, we proved in Lemma~\ref{lem1} that there is almost surely no other combination of three landmarks (excluding the correct one) that can produce the same measurements $\nbd = \left[ d_1, d_2, d_3\right]$, meaning that the correct landmark combination can almost surely be identified if the measurements are noise-free. 
To identify the correct landmark combination when more than two landmarks are involved, we utilize the pairwise geometric constraints between two landmarks, as described previously.
If all pairs of landmarks in a given combination satisfy their corresponding pairwise geometric constraints, we consider this combination as a potential solution and include it in the solution set.
Next, we illustrate the pairwise geometric constraints when three range measurements are available.

\subsection{The Pairwise Geometric Constraints}

\begin{figure} 
	\centering
	\includegraphics[width=0.35\textwidth]{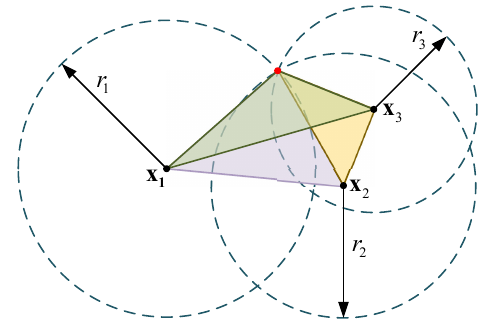}
	\caption{An illustration of the geometric constraints among the target and three landmarks. $r_i$ represents the range measurement of the distance between the target and the $i$-th landmark located at $\nbx_i = (x_i, y_i)$. Blue dashed circles with radii $r_i$ represent the potential locations of the target.}
    \label{fig:3obs}
\end{figure}

%The triangle inequality for a pair of landmarks discussed in section \ref{sec:two-ranges} can be extended by considering all triangles formed by the landmark combination with more than two landmarks.

In~\figref{fig:3obs}, we illustrate the geometric constraints involving three landmarks. 
When two landmarks are selected out of the three, three distinct triangles are formed, each involving the target and two landmarks. 
Each triangle has its corresponding geometric constraints given in Equation~\eqref{eq:tri-inequ-noise}.
Thus, we have six inequalities for three triangles. 
These geometric constraints can be interpreted as follows: if we select two arbitrary landmarks and associate them with measurements $r_1$ and $r_2$, the possible locations for the target are the two intersection points of the circles with $r_1$ and $r_2$, centered at the respective landmarks.
The six inequalities ensure that all three circles intersect at a single point, representing the potential location of the target.
Using this method, we next derive the localizability probability based on these pairwise geometric constraints.

\subsection{The Localizability Probability}

Suppose the target obtains $N$ range measurements $\ncalI = \left\{\left(r_i, m_i\right), i = 1, \dots, N\right\}$ from the correct landmark combination $c^* = \left[\nbx_1, \nbx_2, \cdots, \nbx_N\right]$, where $\nbx_j$ is the location of $i$-th landmark in the combination. 
We use pairwise geometric constraints based on these $N$ measurements to determine the observed landmark combination. 
We check the landmark combinations in $\ncalC$ and add those satisfied the pairwise geometric constraints in the solution set $\ncalS$. 
The same estimation process in Section~\ref{loc-prob} is used to obtain the estimated landmark combination. 
In this case, we derive the localizability probability in the following theorem.

\begin{theorem} \label{the2}
When having $N$ measurements and employing pairwise geometric constraints, the upper bound of the localizability probability is
\begin{equation} \label{eq:the2}
    \nbbP\!\left[ \hat{c} = c^* \mid N \right] \leq \mathbb{E}_{\nbR, \nbM, |\ncalC_n|} \left\{\frac{1 - (1 - p_f)^{|\ncalC_n|}}{|\ncalC_n| p_f} \right\},
\end{equation}
where $\nbR$ and $\nbM$ are random vectors representing the range measurements and their corresponding marks, $|\ncalC_n|$ is the number of elements in the combination set that contains the correct $n$-th landmark located at $\nbx_n$.
\end{theorem}

\begin{IEEEproof}
By definition, the localizability probability is the probability of correctly estimating the landmark combination, which requires accurately identifying all landmarks in the combination. This can be represented as:
\begin{align}
    \nbbP\!\left[ \hat{c} = c^* \mid N \right] =&\, \nbbP\!\left[ \hat{\nbx}_1 = \nbx_1, \cdots, \hat{\nbx}_N = \nbx_N \mid N \right]\\
    \overset{(a)}{=} &\,\nbbP\!\left[\hat{\nbx}_n = \nbx_n \mid N, \hat{c}_{-n} = c_{-n}^*\right] \notag \\
    & \cdot \nbbP\!\left[\hat{c}_{-n} = c_{-n}^* \mid N\right]\\
    \overset{(b)}{\leq}&\, \nbbP\!\left[\hat{\nbx}_n = \nbx_n \mid N, \hat{c}_{-n} = c_{-n}^*\right],
\end{align}
where $\hat{c}_{-n} = \left[\hat{\nbx}_1, \dots, \hat{\nbx}_{n-1}, \hat{\nbx}_{n+1}, \dots, \hat{\nbx}_N\right]$ is the estimated landmark combination excluding the $n$-th entry, $c_{-n}^* = \left[\nbx_1, \dots, \nbx_{n-1}, \nbx_{n+1}, \dots, \nbx_N\right]$ is the correct landmark combination exclude the $n$-th entry, (a) follows from the joint probability formula for dependent events, (b) follows from $P\!\left[\hat{c}_{-n} = c_{-n}^* \mid N \right] \leq 1$. The probability $\nbbP\!\left[\hat{\nbx}_n = \nbx_n \mid N, \hat{c}_{-n} = c_{-n}^*\right]$ represents that, given all other landmarks are correctly estimated, the $n$-th landmark is also correctly estimated.

The $n$-th landmark is estimated by checking the pairwise geometric constraints that involve the $n$-th measurement $\left(r_n, m_n\right)$.
We write the conditional probability of correctly identifying the $n$-th landmark as follows:
\begin{align}
	\nbbP&\!\left[\hat{\nbx}_n = \nbx_n \mid N, \hat{c}_{-n} = c_{-n}^*\right] \notag \\
	=& \mathbb{E}_{\nbR, \nbM, |\ncalC_n|} \!\left\{ \nbbP\!\left[ \hat{\nbx}_n = \nbx_n \mid \nbr, \nbm, |\ncalC_n|, \hat{c}_{-n} = c_{-n}^*\right]\right\}\\
	\leq& \mathbb{E}_{\nbR, \nbM, |\ncalC_n|} \!\left\{ \nbbP\!\left[ \hat{c}_{i,n} = c_{i,n} \mid \nbr, \nbm, |\ncalC_n|,  \hat{c}_{-n} = c^*_{-n} \right] \right\}, \label{eq:one-lmk-error}
\end{align}
where $c_{i, n} = \left[\hat{\nbx}_i, \hat{\nbx}_{n}\right]$ is the landmark pair including the $i$-th and $n$-th entries of the landmark combination, $\hat{c}_{i, n}$ is the estimation of $c_{i, n}$. 
We can see that~\eqref{eq:one-lmk-error} is similar to the expression in~\eqref{eq:loc-prob}, except that~\eqref{eq:one-lmk-error} is conditioned on the assumption that the landmarks $\hat{c}_{-n}$ have already been correctly identified.
In this case, if one of the observed landmark $\nbx_i$ is known, the number of potential landmark pairs $c_{i, n}$ is equivalent to the number of landmarks with mark $m_n$ in the AOI, represented by $|\ncalC_n|$. 
Thus, using the result in Theorem~\ref{the1}, the probability of correctly identifying $c_{i, n}$ can be written as
\begin{align}
	&\mathbb{E}_{\nbR, \nbM, |\ncalC_n|} \!\left\{ \nbbP\!\left[ \hat{c}_{i,n} = c_{i,n} \mid \nbr, \nbm, |\ncalC_n|, \hat{c}_{-n} = c^*_{-n}\right]\right\} \notag\\
	&= \mathbb{E}_{\nbR, \nbM, |\ncalC_n|} \left\{p_t \cdot \frac{1 - \left(1 - p_f\right)^{|\ncalC_n|}}{|\ncalC_n| \cdot p_f} \right\}.
\end{align}
This completes the proof of Theorem~\ref{the2}.
\end{IEEEproof}

Theorem~\ref{the2} provides an analytical upper bound for the localizability probability when utilizing the pairwise geometric constraints. Specifically, the localizability is upper-bounded by the probability of correctly estimating one of the landmarks in the combination, given that all other landmarks have already been correctly identified. 
Increasing the number of measurements improves localizability but does not fully resolve the challenge of uniquely identifying individual landmarks within a combination under the given geometric constraints.
In the next section, we will verify these analytical results and offer additional insights for system design.

\section{Numerical Results}

\begin{figure*}
    \centering
	\subfigure[]{\includegraphics[width=0.32\textwidth]{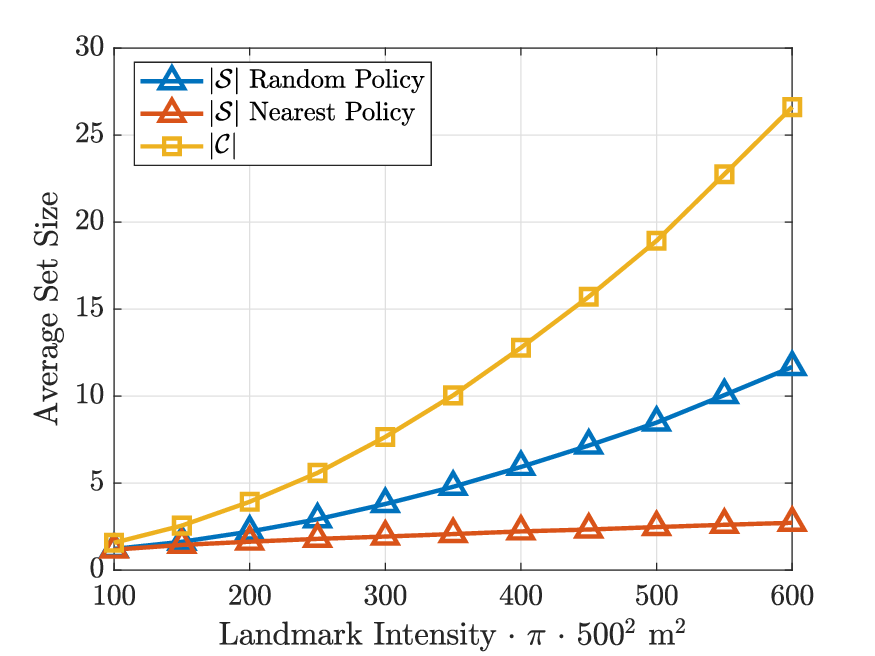}
	\label{fig:set-size}}
	\hfill
	\centering
	\subfigure[]{\includegraphics[width=0.32\textwidth]{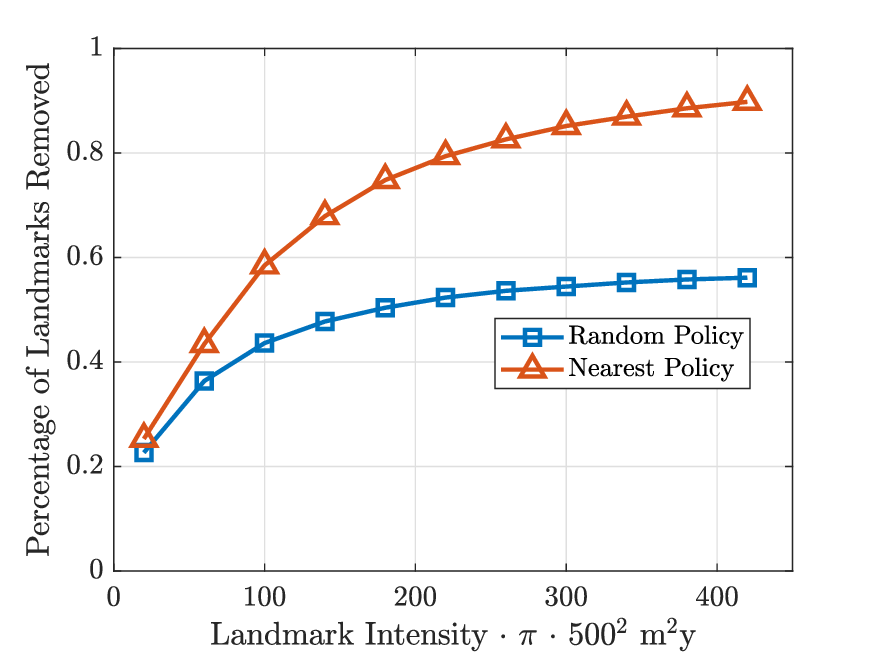}
	\label{fig:reduce}}
	\hfill
    \centering
	\subfigure[]{\includegraphics[width=0.32\textwidth]{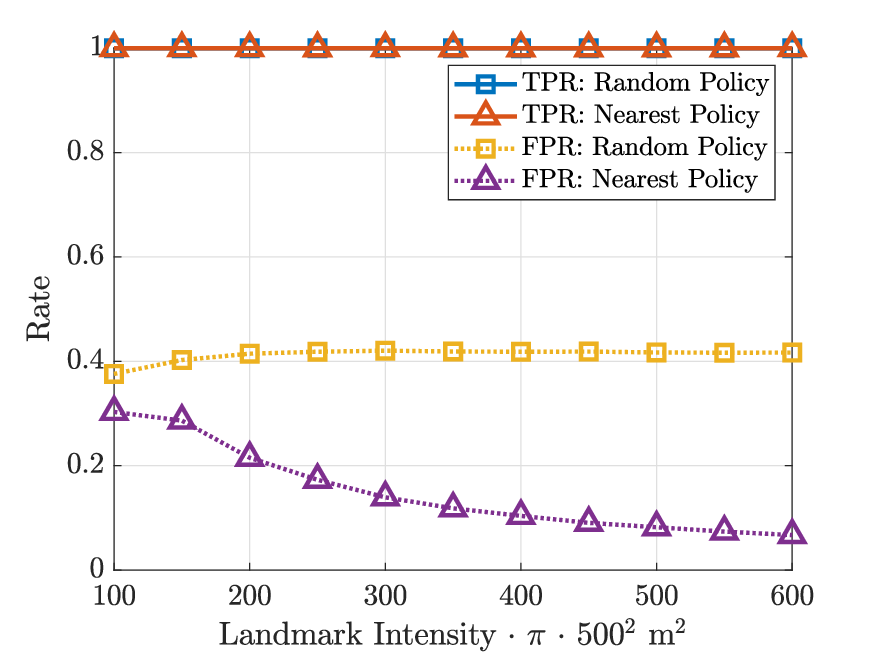}
	\label{fig:rate}}
	\caption{The performance comparison between two observation policies in terms of (a) the sizes of the solution set $\ncalS$, (b) the percentage of landmark combinations removed by the proposed geometric constraints, and (c) the true positive rate and false positive rate evaluated in the stochastic setting.}
\end{figure*}

\begin{figure*}
	\centering
	\subfigure[]{\includegraphics[width=0.32\textwidth]{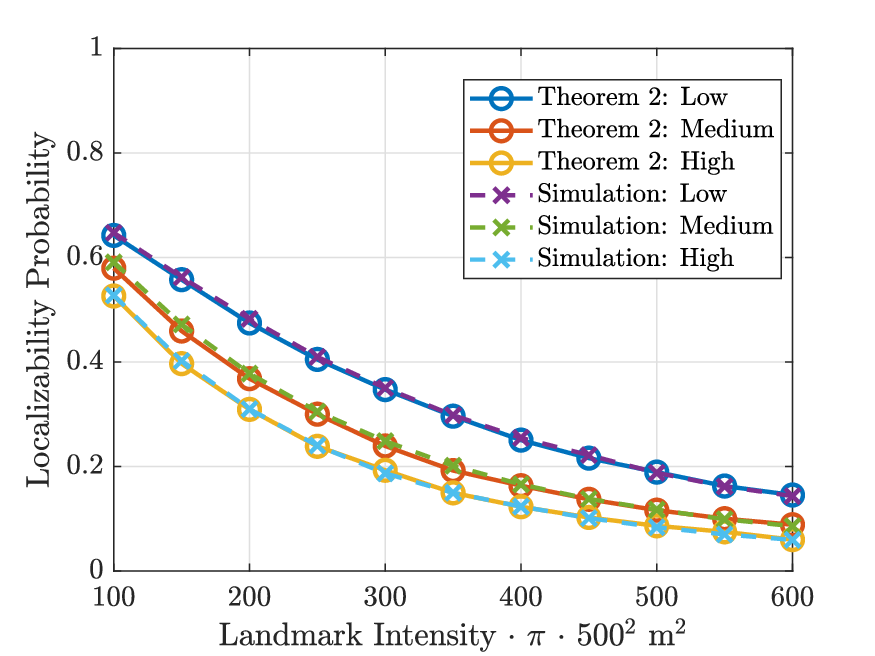}
	\label{fig:pe2}}
	\hfill
	\subfigure[]{\includegraphics[width=0.32\textwidth]{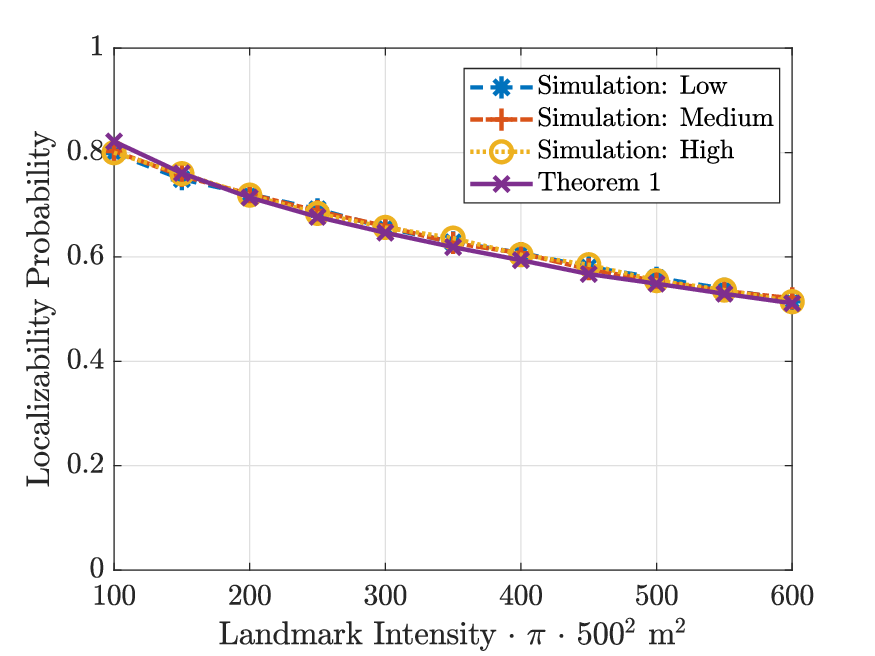}
	\label{fig:pe2-near}}
    \subfigure[]{\includegraphics[width=0.32\textwidth]{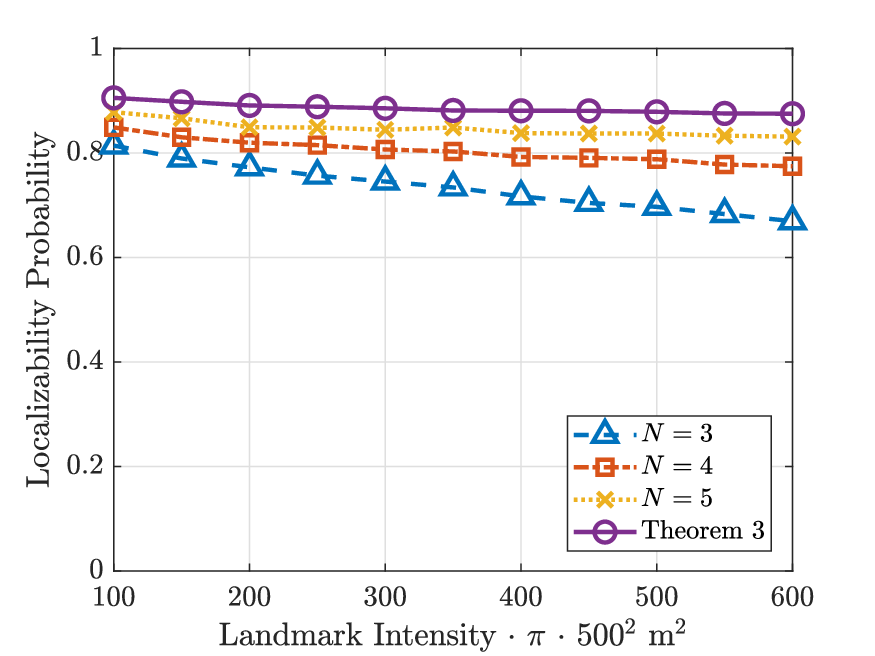}
	\label{fig:pe3}}
	\caption{The localizability probability (a) with two range measurements under the random policy, with $T = 0.2$ and $\sigma = 0.3$, (b) with two range measurements under the nearest policy, with $T = 0.2$ and $\sigma = 0.3$ (c) with $N$ range measurements under the nearest policy, with $T = 0.2$ and $\sigma = 0.3$.}
\end{figure*}

We apply the proposed geometric constraints to identify the observed landmark combination and evaluate the localization performance by comparing the results with the analytically derived expression of the localizability probability. 
We consider the AOI to have the size of a city center with a radius $d_a = 500{\rm m}$.
Within this AOI, $M = 16$ types of landmarks are placed with a uniform intensity $\lambda$. 
The chosen landmark densities are inspired by real-world examples:
(i) New York City with over $16000$ bus stops across an area of $778\,{\rm km}^2$, and (ii) Central Park containing $1863$ lampposts within an area of $863$ acres.
Based on these, the landmark density $\lambda$ in our model varies from $\frac{100}{\pi \cdot 500^2}$ to $\frac{600}{\pi \cdot 500^2}$. 
These landmarks belong to different types with varying visibility distances. We consider three visibility profiles for each type of landmark:
\begin{itemize}
    \item Low visibility: $d_m = 9+m\,$m, resulting in visibility distances ranging from $10\,$m to $25\,$m,
    \item Medium visibility: $d_m = 19+m\,$m, resulting in visibility distances ranging from $20\,$m to $35\,$m,
    \item High visibility: $d_m = 29+m\,$m, resulting in visibility distances ranging from $30\,$m to $45\,$m,
\end{itemize}
where $m \in {1, 2, \dots, M}$ is the index representing the landmark type.
The target is placed randomly within the AOI, and landmarks within the specified visibility distances are considered observable.
The landmark combination is considered to satisfy the geometric constraints if the resulting triangle inequality holds within a threshold tolerance of $T = 0.2$.
The average size of the combination set $\ncalC$, and the solution set $\ncalS$ under both the {\em random policy} and {\em nearest policy} are plotted in~\figref{fig:set-size}. 
We observe that the sizes of combinations sets $|\ncalS|$ are significantly reduced compared to $|\ncalC|$, demonstrating that the geometric constraints efficiently remove landmark combinations.
The percentages of landmark combinations removed by the proposed geometric constraint are presented in~\figref{fig:reduce}.
When comparing the result differences in the two observation policies, the {\em nearest policy} removed more landmark combinations than the {\em random policy}, which means the {\em nearest policy} is more efficient in identifying the correct landmark combinations. 
This suggests that selecting landmarks closer to the target tightens the geometric constraints and improves its localizability.
Next, we plot the corresponding TPR $p_t$ and FPR $p_f$ in~\figref{fig:rate}. These results are derived in Section~\ref{sec:geo-feature}, representing the likelihood of retaining the correct landmark combination and other landmark combinations in the solution set.
We observe that as landmark intensity increases, the false positive rate of the nearest policy decreases while the false positive rate of the random policy remains constant.
This is a direct result of selecting landmarks closer to the target, which tightens the geometric constraints and thus removes more landmark combinations.

We then simulate the localizability probability for both observation policies when having two measurements.
We compare the simulation result with the analytical expressions given in Theorem~\ref{the1} and Theorem~\ref{the3}.
For the random policy, the closed-form expression for the localizability probability is derived in Theorem~\ref{the3}.
The localizability probabilities are illustrated in \figref{fig:pe2} and \figref{fig:pe2-near} for both observation policies.
We observe that the probability of correctly identifying the landmark combinations aligns closely with the theoretical predictions. 
This validation verifies the accuracy of the analytical expressions for the localizability probability derived for both observation policies.
Furthermore, we observe that an increase in landmark density results in a decrease in localizability probability. The larger number of landmarks leads to more potential combinations, making accurate estimation of the observed landmark combination more difficult.
Additionally, the nearest policy has higher localization probabilities than the random policy, since the selected landmarks offer tighter geometric constraints.

We also simulate the probability of correctly identifying the landmark combination using more than two range measurements.
The simulation result is compared against the theoretical upper bound derived in Theorem~\ref{the2}. 
As shown in~\figref{fig:pe3}, the localizability probability increases with the number of range measurements $N$, indicating a greater chance of correctly identifying the landmark combination with additional measurements.
However, this improvement is constrained by the upper bound established in Theorem~\ref{the2}.
Even with additional measurements, identifying an individual landmark within the landmark combination remains challenging with the proposed geometric constraints.
This finding emphasizes that although increasing the number of measurements enhances localizability, the improvement is inherently limited.
These results provide valuable metrics for evaluating the feasibility of landmark-based localization in various environments.

\section{Conclusions}
This paper introduces a novel stochastic geometry-based framework for vision-based localization. Since landmarks detected in vision data may not always be uniquely identifiable, we model them as a marked PPP. We assume that landmarks are visible when they fall within a defined visibility range. Vision sensors measure the distances to these landmarks, and we use these measurements to establish geometric constraints that help identify the correct combination of landmarks associated with the measurements. We derive the probability of correctly identifying the landmark combination to evaluate the localization accuracy.
This work is the first to tackle the challenge of landmark non-uniqueness in vision-based localization and to analyze their properties within this framework.

Given the novel framework presented in this paper, there are several potential directions for future research. Two specific extensions include (i) exploring additional geometric constraints in the analysis, (ii) developing more observation policies for selecting a subset of landmarks from a larger set of visible landmarks, and (iii) deriving the analytical localizability probability when more than two measurements are available.
Overall, this paper bridges the fields of stochastic geometry, localization, and computer vision, potentially opening up new avenues for research in these areas.

\begin{comment}
This paper presents a new stochastic geometry-based framework for vision-based localization. Since landmarks appearing in vision data may not be uniquely identifiable, we modeled them as a marked PPP.
Further, we assume that landmarks are visible if their ranges to the target are within the visibility distance.
Vision sensors measure the distances to landmarks, and we use these measurements to construct geometric constraints to identify the correct landmark combination from which measurements are obtained.
We derived the probability of correctly identifying the landmark combination to evaluate the localization performance.
This work is the first to address the non-uniqueness of landmarks and analyze their properties.

Given the novel setting of this paper, this work can be extended in multiple directions. Two extensions that are specific to this paper are (i) exploring other geometric constraints in the analysis, (ii) incorporating more observation policies for selecting a limited set of observations from a larger set of visible landmarks, and (iii) deriving the analytical localizability probability for more than two measurements. 
Overall, this paper connects stochastic geometry, localization, and computer vision, which could inspire a new direction of investigation. 
\end{comment}

\bibliographystyle{IEEEtran}
\bibliography{citation}

% Generated by IEEEtran.bst, version: 1.14 (2015/08/26)
\begin{thebibliography}{10}
\providecommand{\url}[1]{#1}
\csname url@samestyle\endcsname
\providecommand{\newblock}{\relax}
\providecommand{\bibinfo}[2]{#2}
\providecommand{\BIBentrySTDinterwordspacing}{\spaceskip=0pt\relax}
\providecommand{\BIBentryALTinterwordstretchfactor}{4}
\providecommand{\BIBentryALTinterwordspacing}{\spaceskip=\fontdimen2\font plus
\BIBentryALTinterwordstretchfactor\fontdimen3\font minus
  \fontdimen4\font\relax}
\providecommand{\BIBforeignlanguage}[2]{{%
\expandafter\ifx\csname l@#1\endcsname\relax
\typeout{** WARNING: IEEEtran.bst: No hyphenation pattern has been}%
\typeout{** loaded for the language `#1'. Using the pattern for}%
\typeout{** the default language instead.}%
\else
\language=\csname l@#1\endcsname
\fi
#2}}
\providecommand{\BIBdecl}{\relax}
\BIBdecl

\bibitem{conf1}
H.~Hu, H.~S. Dhillon, and R.~M. Buehrer, ``Landmark-based localization using
  range measurements: A stochastic geometry perspective,'' in \emph{IEEE/IFIP
  WiOpt Workshops}.\hskip 1em plus 0.5em minus 0.4em\relax IEEE, May 2023, pp.
  425--432.

\bibitem{wireless}
C.~Yang, S.~Mao, and X.~Wang, ``An overview of {3GPP} positioning standards,''
  \emph{GetMobile: Mobile Computing and Commun.}, vol.~26, no.~1, pp. 9--13,
  2022.

\bibitem{crlb1}
T.~Jia and R.~M. Buehrer, ``A new {Cramer-Rao} lower bound for {TOA-based}
  localization,'' in \emph{Proc., IEEE MILCOM}, Nov. 2008, pp. 1--5.

\bibitem{crlb2}
L.~Gui, M.~Yang, H.~Yu, J.~Li, F.~Shu, and F.~Xiao, ``A {Cramer-Rao} lower
  bound of {CSI-based} indoor localization,'' \emph{IEEE Trans. on Veh.
  Technology}, vol.~67, no.~3, pp. 2814--2818, 2017.

\bibitem{crlb3}
B.~Huang, L.~Xie, and Z.~Yang, ``{TDOA-based} source localization with
  distance-dependent noises,'' \emph{IEEE Trans. on Wireless Commun.}, vol.~14,
  no.~1, pp. 468--480, 2014.

\bibitem{del2017survey}
J.~A. del Peral-Rosado, R.~Raulefs, J.~A. L{\'o}pez-Salcedo, and
  G.~Seco-Granados, ``Survey of cellular mobile radio localization methods:
  From {1G} to {5G},'' \emph{IEEE Commun. Surveys \& Tutorials}, vol.~20,
  no.~2, pp. 1124--1148, 2017.

\bibitem{buehrerhandbook}
R.~Zekavat and R.~M. Buehrer, \emph{Handbook of position location: Theory,
  practice and advances}.\hskip 1em plus 0.5em minus 0.4em\relax John Wiley \&
  Sons, 2011, vol.~27.

\bibitem{city-scale}
G.~Schindler, M.~Brown, and R.~Szeliski, ``City-scale location recognition,''
  in \emph{Proc., IEEE Conf. on Computer Vision and Pattern Recognition
  (CVPR)}, Jun. 2007.

\bibitem{visual-place}
A.~Torii, J.~Sivic, T.~Pajdla, and M.~Okutomi, ``Visual place recognition with
  repetitive structures,'' in \emph{Proc., IEEE Conf. on Computer Vision and
  Pattern Recognition (CVPR)}, Jun. 2013.

\bibitem{lin2013cross}
T.-Y. Lin, S.~Belongie, and J.~Hays, ``Cross-view image geolocalization,'' in
  \emph{Proc., IEEE Conf. on Computer Vision and Pattern Recognition (CVPR)},
  Jun. 2013, pp. 891--898.

\bibitem{singh2016semantically}
G.~Singh and J.~Ko{\v{s}}eck{\'a}, ``Semantically guided geo-location and
  modeling in urban environments,'' \emph{Large-Scale Visual Geo-Localization},
  pp. 101--120, 2016.

\bibitem{crandall2016recognizing}
D.~J. Crandall, Y.~Li, S.~Lee, and D.~P. Huttenlocher, ``Recognizing landmarks
  in large-scale social image collections,'' \emph{Large-Scale Visual
  Geo-Localization}, pp. 121--144, 2016.

\bibitem{csurka2004visual}
G.~Csurka, C.~Dance, L.~Fan, J.~Willamowski, and C.~Bray, ``Visual
  categorization with bags of keypoints,'' in \emph{European Conference on
  Computer Vision (ECCV)}, vol.~1, no. 1-22.\hskip 1em plus 0.5em minus
  0.4em\relax Prague, May 2004, pp. 1--2.

\bibitem{li2016worldwide}
Y.~Li, N.~Snavely, D.~P. Huttenlocher, and P.~Fua, ``Worldwide pose estimation
  using {3D} point clouds,'' \emph{Large-Scale Visual Geo-Localization}, pp.
  147--163, 2016.

\bibitem{saurer2016image}
O.~Saurer, G.~Baatz, K.~K{\"o}ser, L.~Ladick{\`y}, and M.~Pollefeys,
  ``Image-based large-scale geo-localization in mountainous regions,'' in
  \emph{Large-Scale Visual Geo-Localization}.\hskip 1em plus 0.5em minus
  0.4em\relax Springer, 2016, pp. 205--223.

\bibitem{ground2aerial}
S.~Verde, T.~Resek, S.~Milani, and A.~Rocha, ``Ground-to-aerial viewpoint
  localization via landmark graphs matching,'' \emph{IEEE Signal Processing
  Letters}, vol.~27, pp. 1490--1494, 2020.

\bibitem{kendall2015posenet}
A.~Kendall, M.~Grimes, and R.~Cipolla, ``Posenet: A convolutional network for
  real-time 6-{DOF} camera relocalization,'' in \emph{Proc., IEEE Intl. Conf.
  on Computer Vision (ICCV)}, Dec. 2015, pp. 2938--2946.

\bibitem{geometry-aware}
S.~Brahmbhatt, J.~Gu, K.~Kim, J.~Hays, and J.~Kautz, ``Geometry-aware learning
  of maps for camera localization,'' in \emph{Proc., IEEE Conf. on Computer
  Vision and Pattern Recognition (CVPR)}, Jun. 2018, pp. 2616--2625.

\bibitem{ding2019camnet}
M.~Ding, Z.~Wang, J.~Sun, J.~Shi, and P.~Luo, ``{CamNet}: {Coarse-to-fine}
  retrieval for camera re-localization,'' in \emph{Proc., IEEE Intl. Conf. on
  Computer Vision (ICCV)}, Oct. 2019, pp. 2871--2880.

\bibitem{walch2017image}
F.~Walch, C.~Hazirbas, L.~Leal-Taixe, T.~Sattler, S.~Hilsenbeck, and
  D.~Cremers, ``Image-based localization using {LSTMs} for structured feature
  correlation,'' in \emph{Proc., IEEE Intl. Conf. on Computer Vision (ICCV)},
  Oct. 2017, pp. 627--637.

\bibitem{7scene}
J.~Shotton, B.~Glocker, C.~Zach, S.~Izadi, A.~Criminisi, and A.~Fitzgibbon,
  ``Scene coordinate regression forests for camera relocalization in {RGB-D}
  images,'' in \emph{Proc., IEEE Conf. on Computer Vision and Pattern
  Recognition (CVPR)}, Jun. 2013, pp. 2930--2937.

\bibitem{orb-slam}
R.~Mur-Artal, J.~M.~M. Montiel, and J.~D. Tardos, ``{ORB-SLAM}: a versatile and
  accurate monocular {SLAM} system,'' \emph{IEEE Trans. on Robotics}, vol.~31,
  no.~5, pp. 1147--1163, 2015.

\bibitem{lsd-slam}
J.~Engel, T.~Sch{\"o}ps, and D.~Cremers, ``{LSD-SLAM}: Large-scale direct
  monocular {SLAM},'' in \emph{European Conference on Computer Vision
  (ECCV)}.\hskip 1em plus 0.5em minus 0.4em\relax Springer, Sep. 2014, pp.
  834--849.

\bibitem{dtam}
R.~A. Newcombe, S.~J. Lovegrove, and A.~J. Davison, ``{DTAM}: Dense tracking
  and mapping in real-time,'' in \emph{Proc., IEEE Intl. Conf. on Computer
  Vision (ICCV)}.\hskip 1em plus 0.5em minus 0.4em\relax IEEE, Nov. 2011, pp.
  2320--2327.

\bibitem{chen2022overview}
W.~Chen, G.~Shang, A.~Ji, C.~Zhou, X.~Wang, C.~Xu, Z.~Li, and K.~Hu, ``An
  overview on visual {SLAM}: From tradition to semantic,'' \emph{Remote
  Sensing}, vol.~14, no.~13, p. 3010, 2022.

\bibitem{talukder2003real}
A.~Talukder, S.~Goldberg, L.~Matthies, and A.~Ansar, ``Real-time detection of
  moving objects in a dynamic scene from moving robotic vehicles,'' in
  \emph{IEEE/RSJ Intl. Conf. on Intelligent Robots and Systems (IROS)},
  vol.~2.\hskip 1em plus 0.5em minus 0.4em\relax IEEE, Oct. 2003, pp.
  1308--1313.

\bibitem{dornhege2006visual}
C.~Dornhege and A.~Kleiner, ``Visual odometry for tracked vehicles,'' in
  \emph{In Proc. of the IEEE Int. Workshop on Safety, Security and Rescue
  Robotics (SSRR)}, 2006.

\bibitem{zhang2009visual}
T.~Zhang, X.~Liu, K.~K{\"u}hnlenz, and M.~Buss, ``Visual odometry for the
  autonomous city explorer,'' in \emph{IEEE/RSJ Intl. Conf. on Intelligent
  Robots and Systems (IROS)}.\hskip 1em plus 0.5em minus 0.4em\relax IEEE, Oct.
  2009, pp. 3513--3518.

\bibitem{civera2011towards}
J.~Civera, D.~G{\'a}lvez-L{\'o}pez, L.~Riazuelo, J.~D. Tard{\'o}s, and J.~M.~M.
  Montiel, ``Towards semantic {SLAM} using a monocular camera,'' in
  \emph{IEEE/RSJ Intl. Conf. on Intelligent Robots and Systems (IROS)}, Sep.
  2011.

\bibitem{salas2013slam++}
R.~F. Salas-Moreno, R.~A. Newcombe, H.~Strasdat, P.~H. Kelly, and A.~J.
  Davison, ``{SLAM}++: Simultaneous localisation and mapping at the level of
  objects,'' in \emph{Proc., IEEE Conf. on Computer Vision and Pattern
  Recognition (CVPR)}, Jun. 2013, pp. 1352--1359.

\bibitem{engel2014lsd}
J.~Engel, T.~Sch{\"o}ps, and D.~Cremers, ``{LSD-SLAM}: Large-scale direct
  monocular {SLAM},'' in \emph{European Conference on Computer Vision (ECCV)},
  Sep. 2014, pp. 834--849.

\bibitem{newcombe2011dtam}
R.~A. Newcombe, S.~J. Lovegrove, and A.~J. Davison, ``{DTAM}: {Dense} tracking
  and mapping in real-time,'' in \emph{Proc., IEEE Intl. Conf. on Computer
  Vision (ICCV)}, Nov. 2011, pp. 2320--2327.

\bibitem{sattler2011fast}
T.~Sattler, B.~Leibe, and L.~Kobbelt, ``Fast image-based localization using
  direct {2D-to-3D} matching,'' in \emph{Proc., IEEE Intl. Conf. on Computer
  Vision (ICCV)}, Nov. 2011, pp. 667--674.

\bibitem{huang2007convergence}
S.~Huang and G.~Dissanayake, ``Convergence and consistency analysis for
  extended kalman filter based {SLAM},'' \emph{IEEE Trans. on Robotics},
  vol.~23, no.~5, pp. 1036--1049, 2007.

\bibitem{valencia2018mapping}
R.~Valencia and J.~Andrade-Cetto, \emph{Mapping, planning and exploration with
  {Pose SLAM}}.\hskip 1em plus 0.5em minus 0.4em\relax Springer, 2018, vol.~74.

\bibitem{ulrich2000appearance}
I.~Ulrich and I.~Nourbakhsh, ``Appearance-based place recognition for
  topological localization,'' in \emph{Proc., IEEE Intl. Conf. on Robotics and
  Automation (ICRA)}, vol.~2.\hskip 1em plus 0.5em minus 0.4em\relax IEEE, Apr.
  2000, pp. 1023--1029.

\bibitem{galvez2012bags}
D.~G{\'a}lvez-L{\'o}pez and J.~D. Tardos, ``Bags of binary words for fast place
  recognition in image sequences,'' \emph{IEEE Trans. on Robotics}, vol.~28,
  no.~5, pp. 1188--1197, 2012.

\bibitem{censi2009achievable}
A.~Censi, ``On achievable accuracy for pose tracking,'' in \emph{Proc., IEEE
  Intl. Conf. on Robotics and Automation (ICRA)}.\hskip 1em plus 0.5em minus
  0.4em\relax IEEE, May 2009, pp. 1--7.

\bibitem{rohde2016localization}
J.~Rohde, J.~E. Stellet, H.~Mielenz, and J.~M. Z{\"o}llner, ``Localization
  accuracy estimation with application to perception design,'' in \emph{Proc.,
  IEEE Intl. Conf. on Robotics and Automation (ICRA)}.\hskip 1em plus 0.5em
  minus 0.4em\relax IEEE, May 2016, pp. 4777--4783.

\bibitem{J2}
H.~Hu, H.~S. Dhillon, and R.~M. Buehrer, ``Foundations of vision-based
  localization: A new approach to localizability analysis using stochastic
  geometry,'' in \emph{arXiv:2409.09525}, 2024.

\bibitem{stochasticbook}
M.~Haenggi, \emph{Stochastic geometry for wireless networks}.\hskip 1em plus
  0.5em minus 0.4em\relax Cambridge University Press, 2012.

\bibitem{cellurstochasticbook}
J.~G. Andrews, A.~K. Gupta, A.~M. Alammouri, and H.~S. Dhillon, \emph{An
  Introduction to Cellular Network Analysis using Stochastic Geometry}.\hskip
  1em plus 0.5em minus 0.4em\relax Morgan \& Claypool Publishers, 2022.

\bibitem{eigen2014depth}
D.~Eigen, C.~Puhrsch, and R.~Fergus, ``Depth map prediction from a single image
  using a multi-scale deep network,'' \emph{Advances in Neural Information
  Processing Systems (NIPS)}, vol.~27, Dec. 2014.

\bibitem{brosh2019accurate}
E.~Brosh, M.~Friedmann, I.~Kadar, L.~Yitzhak~Lavy, E.~Levi, S.~Rippa,
  Y.~Lempert, B.~Fernandez-Ruiz, R.~Herzig, and T.~Darrell, ``Accurate visual
  localization for automotive applications,'' in \emph{Proc., IEEE Conf. on
  Computer Vision and Pattern Recognition (CVPR)}, Jun. 2019.

\bibitem{zamir2010accurate}
A.~R. Zamir and M.~Shah, ``Accurate image localization based on {Google} {Maps}
  street view,'' in \emph{European Conference on Computer Vision (ECCV)}.\hskip
  1em plus 0.5em minus 0.4em\relax Springer, Sep. 2010, pp. 255--268.

\bibitem{distanceDistribution}
S.~Lellouche and M.~Souris, ``Distribution of distances between elements in a
  compact set,'' \emph{Stats}, vol.~3, no.~1, pp. 1--15, 2019.

\end{thebibliography}

\end{document}